\documentclass[sigconf]{acmart}
\usepackage{subcaption}
\usepackage{listings}
\usepackage{framed}
\AtBeginDocument{%
  }

\renewenvironment{leftbar}[1][\hsize]
{%
    \MakeFramed{\hsize#1\advance\hsize-\width\FrameRestore}%
    \raggedright
}
{\endMakeFramed}
\colorlet{punct}{red!60!black}
\definecolor{background}{HTML}{EEEEEE}
\definecolor{delim}{RGB}{20,105,176}
\colorlet{numb}{magenta!60!black}

\lstdefinelanguage{json}{
    basicstyle=\footnotesize\ttfamily,
    stepnumber=1,
    showstringspaces=false,
    breaklines=true,
    frame=lines,
    backgroundcolor=\color{background},
    literate=
     *{0}{{{\color{numb}0}}}{1}
      {1}{{{\color{numb}1}}}{1}
      {2}{{{\color{numb}2}}}{1}
      {3}{{{\color{numb}3}}}{1}
      {4}{{{\color{numb}4}}}{1}
      {5}{{{\color{numb}5}}}{1}
      {6}{{{\color{numb}6}}}{1}
      {7}{{{\color{numb}7}}}{1}
      {8}{{{\color{numb}8}}}{1}
      {9}{{{\color{numb}9}}}{1}
      {:}{{{\color{punct}{:}}}}{1}
      {,}{{{\color{punct}{,}}}}{1}
      {\{}{{{\color{delim}{\{}}}}{1}
      {\}}{{{\color{delim}{\}}}}}{1}
      {[}{{{\color{delim}{[}}}}{1}
      {]}{{{\color{delim}{]}}}}{1},
}

\setcopyright{acmlicensed}
\copyrightyear{2025}
\acmYear{2025}
\acmDOI{} % XXXXXXX.XXXXXXX
\acmConference[]{}%[Conference acronym 'XX]{Make sure to enter the correct
\acmISBN{978-1-4503-XXXX-X/2018/06}

\begin{document}

%%
%% The "title" command has an optional parameter,
%% allowing the author to define a "short title" to be used in page headers.
\title{From LLM-Driven Trading Card Generation to Procedural Relatedness: A Pokémon Case Study}

%%
%% The "author" command and its associated commands are used to define
%% the authors and their affiliations.
%% Of note is the shared affiliation of the first two authors, and the
%% "authornote" and "authornotemark" commands
%% used to denote shared contribution to the research.
\author{Johannes Pfau}
\email{j.pfau@uu.nl}
\orcid{0000-0002-8760-5023}
\affiliation{%
  \institution{Utrecht University}
  \city{Utrecht}
  \country{Netherlands}
}

\author{Panagiotis Vrettis}
\email{p.vrettis@uu.nl}
\orcid{0009-0005-2545-4791}
\affiliation{%
  \institution{Utrecht University}
  \city{Utrecht}
  \country{Netherlands}
}

%%
%% By default, the full list of authors will be used in the page
%% headers. Often, this list is too long, and will overlap
%% other information printed in the page headers. This command allows
%% the author to define a more concise list
%% of authors' names for this purpose.
\renewcommand{\shortauthors}{Pfau \& Vrettis}

%%
%% The abstract is a short summary of the work to be presented in the
%% article.
\begin{abstract}
Since the dawn of Trading Card Games, the genre has grown into a multi-billion-dollar industry engaging millions of analog and digital players worldwide. Popular TCGs rely on regular updates, balance adjustments, and rotating constraints to sustain engagement. Yet, as metagames stabilize, predictable strategies dominate and viable card options diminish, often resulting in repetitive and impaired player experiences.

This paper investigates the use of Large Language Models and Image Diffusion Models for Procedural Content Generation of TCG cards, addressing these challenges by enabling a personalized infinity of card designs. Modern generative AI not only enables large-scale content creation but could even introduce \textit{procedural relatedness}, fostering unique connections between players and their cards. We present a pipeline combining player-centric co-creation, fine-tuned embeddings, local LLMs, and Diffusion Models to generate dynamic, personalized cards while potentially expanding creative range.

We evaluated the pipeline in a user study with 49 participants who generated 196 Pokémon card samples. Participants rated aesthetics and representativeness of visuals and mechanics, and provided qualitative feedback. Results show high satisfaction and indicate that most participants successfully realized their own ideas through prompt adjustments. These findings lay groundwork for future content generation systems and alternatives to conventional metagame evolution through \textit{procedural relatedness}.
\end{abstract}

%%
%% The code below is generated by the tool at http://dl.acm.org/ccs.cfm.
%% Please copy and paste the code instead of the example below.
%%


%%
%% Keywords. The author(s) should pick words that accurately describe
%% the work being presented. Separate the keywords with commas.
%% A "teaser" image appears between the author and affiliation
%% information and the body of the document, and typically spans the
%% page.
\begin{teaserfigure}
  \includegraphics[width=.168\linewidth]{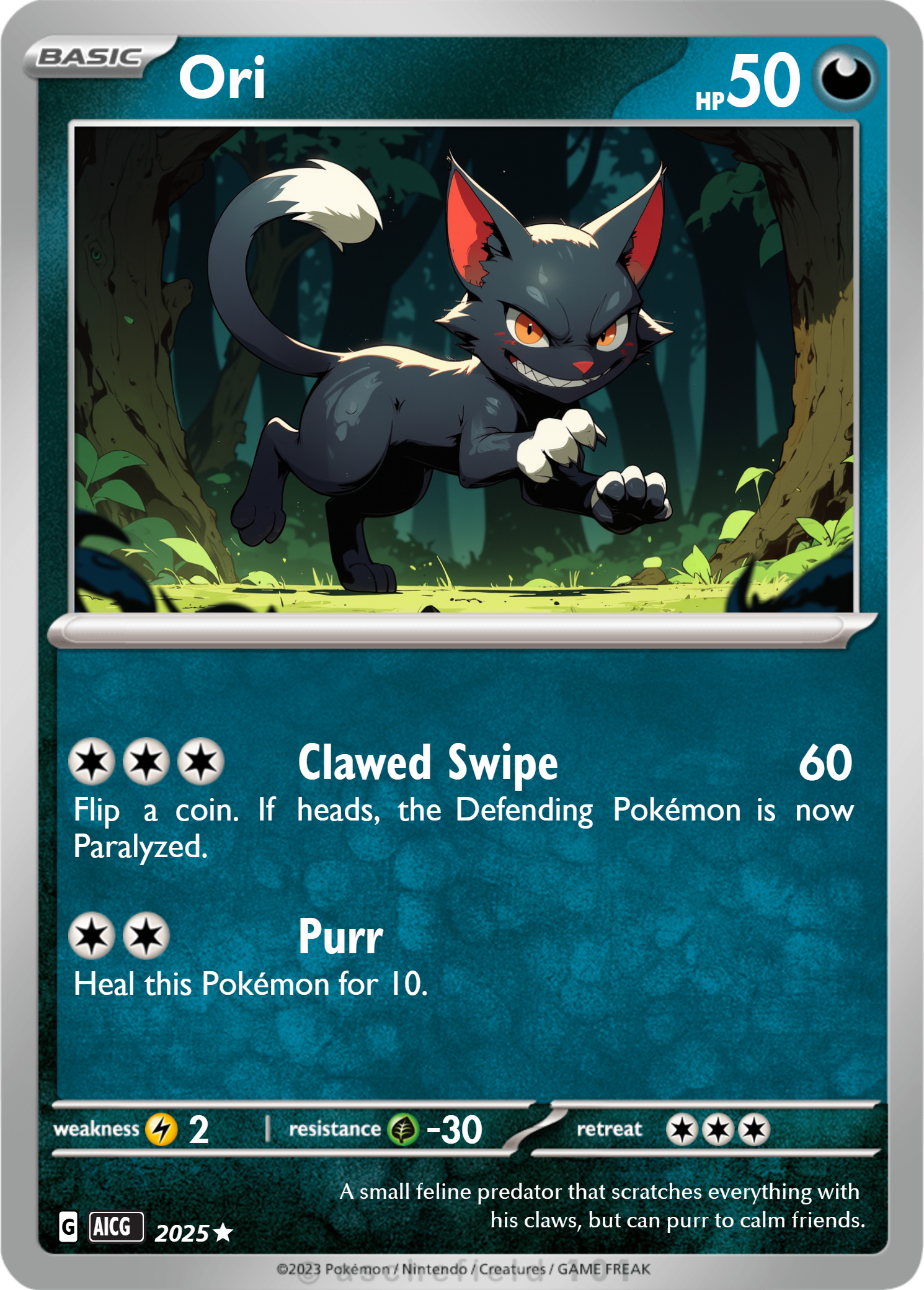}\includegraphics[width=.168\linewidth]{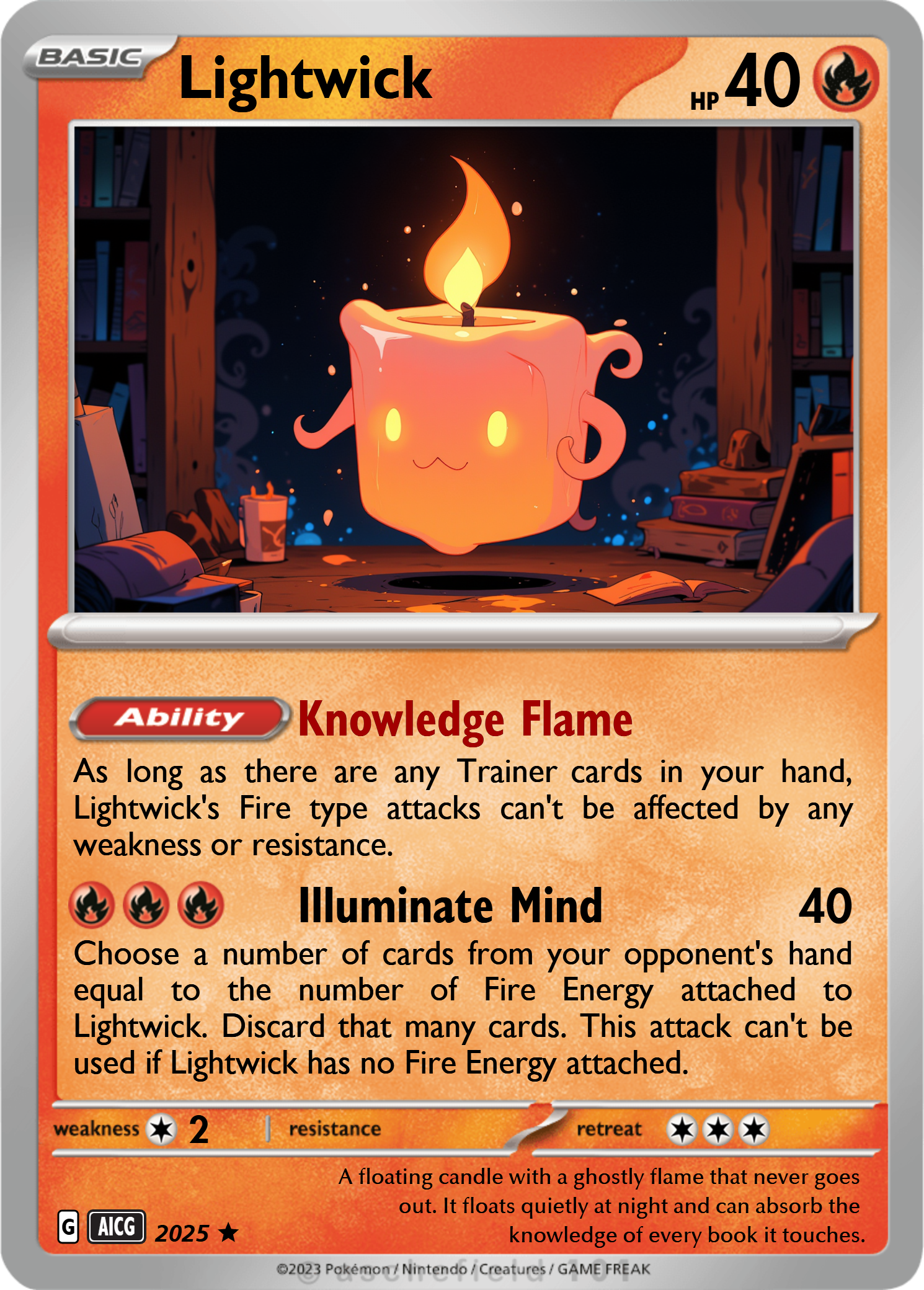}\includegraphics[width=.168\linewidth]{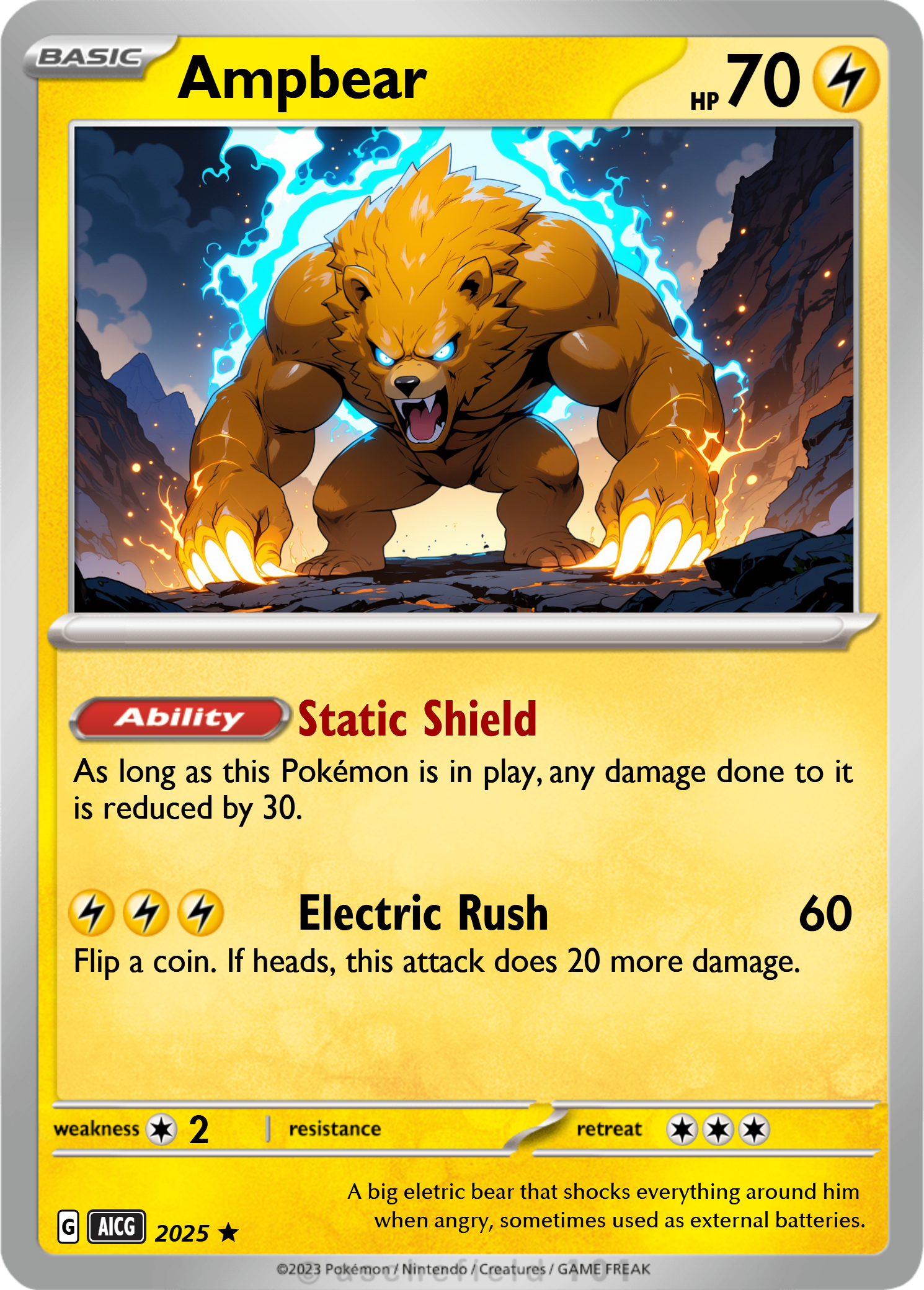}\includegraphics[width=.168\linewidth]{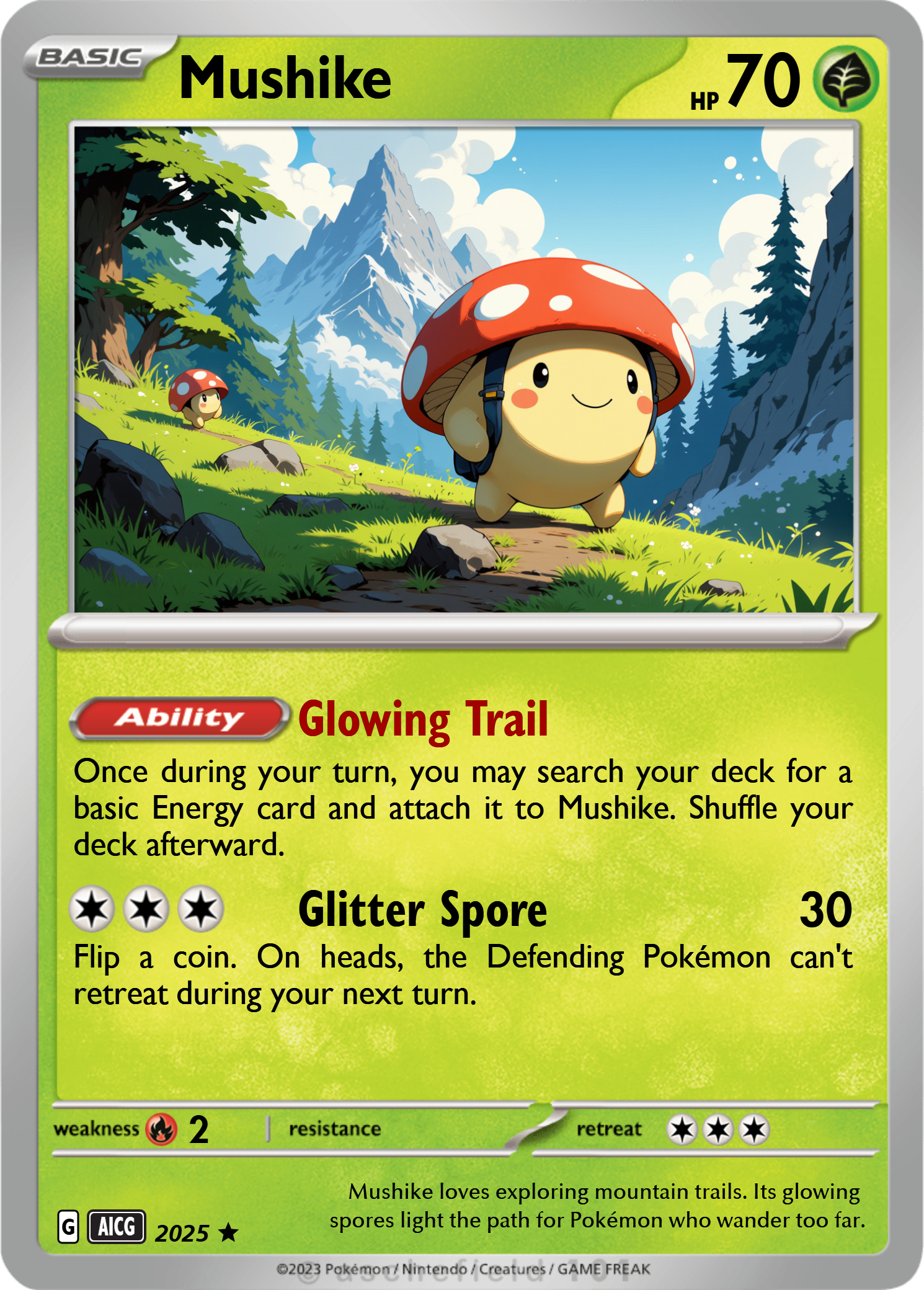}\includegraphics[width=.168\linewidth]{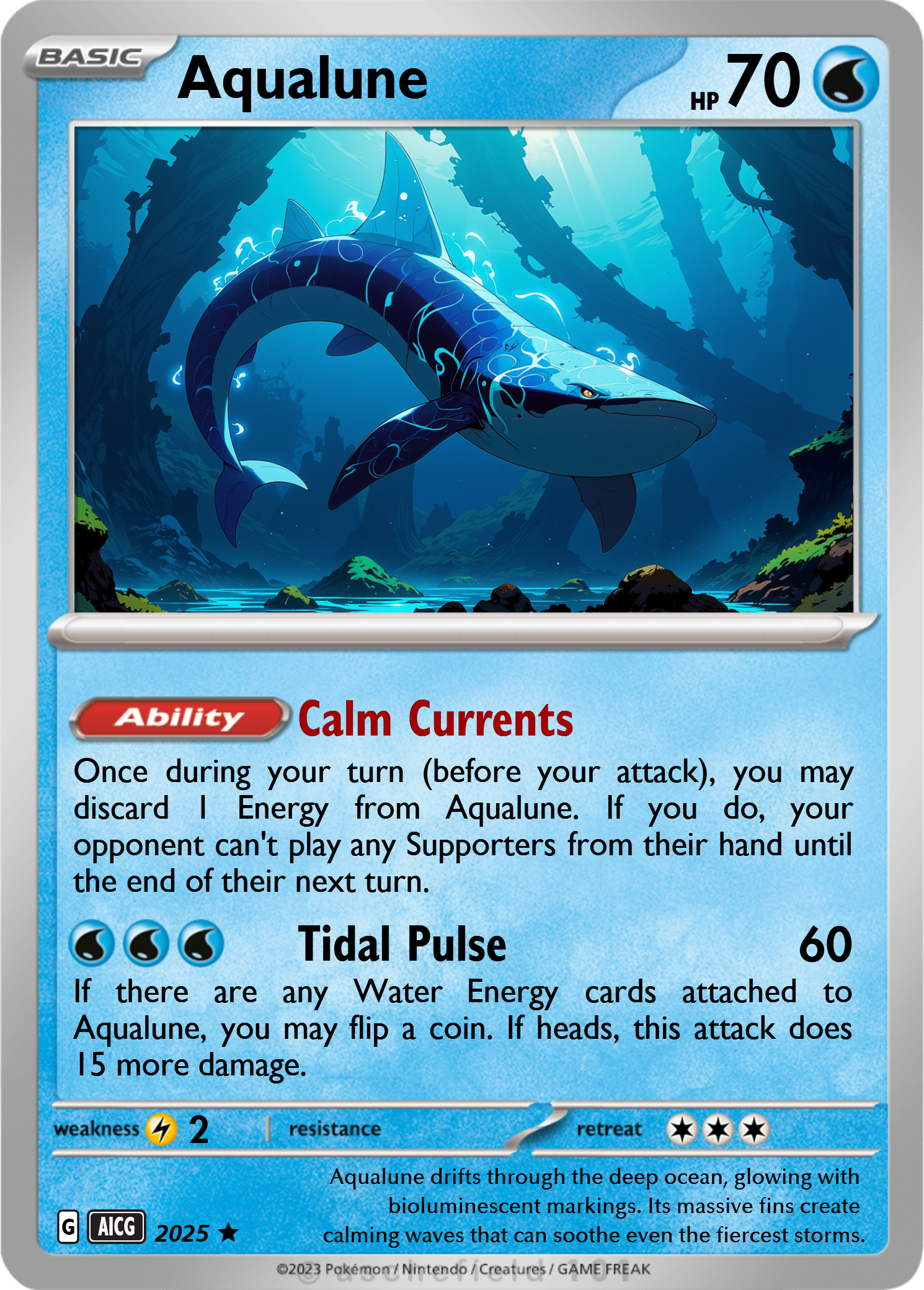}\includegraphics[width=.168\linewidth]{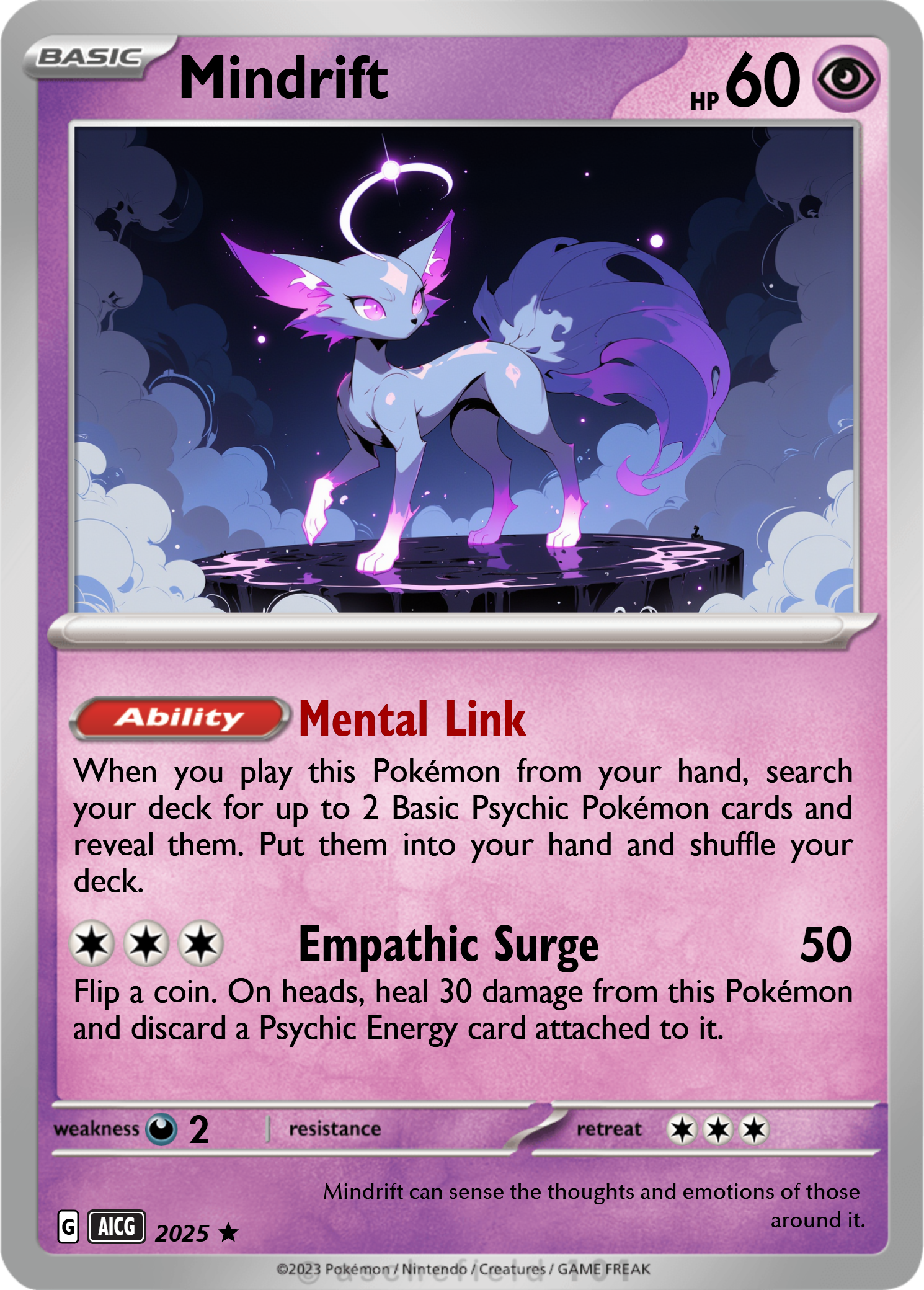}
  \caption{Six sample outputs of the pipeline deployed in this work. From only a name, type, and description \textit{(``Pokédex entry'')}, the model generates abilities, attacks, various stats, and a visual representation.}
  \Description{Six generated Pokémon cards using the pipeline deployed in this work, each showing name, type, description, abilities, and attacks, as well as a visual representation of the Pokémon.}
  \label{fig:teaser}
\end{teaserfigure}

\received{20 February 2007}
\received[revised]{12 March 2009}
\received[accepted]{5 June 2009}

%%
%% This command processes the author and affiliation and title
%% information and builds the first part of the formatted document.
\maketitle

\section{Introduction}
Since their introduction in 1993 with \textit{Magic: The Gathering} \cite{garfield1993magic}, Trading Card Games (TCG) have manifested themselves as a multi-billion dollar industry with steady predicted growth and players of analog and digital versions all over the world \cite{david2010trading, deaux2019move}. \textit{Magic} alone is estimated to engage over 50 million players, has surpassed one billion dollars in revenue, and continues to publish new sets of mostly novel cards every few months \cite{hasbro2024}. 
As with other popular representatives of the genre (such as \textit{Hearthstone} \cite{hearthstone}, \textit{Legends of Runeterra} \cite{games2020legends}, \textit{Pokémon TCG} \cite{pokemonTCG} or \textit{Yu-Gi-Oh!} \cite{yugioh}), \textit{Magic: The Gathering Arena} \cite{MTGA} also secures large numbers of players into recurring online play, to challenge friends, compete for ranks, and/or extend players' collections of cards. Naturally, players are hungry for content and variety in play --- which game developers are happy to contribute to with constantly updated sets, game balance changes, and rotating formats. However, assuming that players are able to overcome the phase of acquiring the most promising or necessary cards of the season, the \textit{metagame} gradually shifts into hard constellations (for example: appearing trends of \textit{aggro} versus \textit{control} versus \textit{combo} decks that soft-dominate each other in a \textit{Rock-Paper-Scissors} fashion) in almost any game and point in time \cite{carter2012metagames,kokkinakis2021metagaming}. The fact that a majority of choosable options become non-desirable over time alone is reason enough to research for alternative directions --- especially, as the resulting metagame can get very limiting, stale, and repetitive. In this paper, we are looking forward to exploring the realm of procedural trading card generation, given its many interesting prospects. For one, echoing the traditional perk of procedural content generation (PCG), it could yield an endless variety of choices, where (ideally) every experience is fresh and interesting anew. Second, a large aspect of TCGs has always been the \textit{trading} aspect, whereas right now, in-game economies are necessarily largely shifted towards very few and very pricey cards, leaving players with large heaps of undesirable (digital or paper) waste, which calls for an approach that yields diverse-enough but viable card sets.
Being under constant pressure to deliver manually created content remains also a steady burden for developers, which can compromise the quality and novelty of newly introduced cards -- often only compensated by releasing slightly stronger cards than before to still keep new cards attractive, inevitably leading to \textit{power creep}, which happens to as good as every live game \cite{mannstrom2022power,magruder2022conservative}. Our main motivation for creating novel cards from scratch is however less often argued: procedural cards could bring back \textit{relatedness} to TCGs, re-instantiating ownership and bonds between cards and players. To express this idea in a more applicatory sense, the most popular Pokémon, \textit{Pikachu}, became that prominent because it was \textit{the} signature Pokémon of the protagonist \textit{Ash Ketchum}, the one he shared all of his adventures with --- not just ``one of many'' Pokémon in his collection, and certainly not the strongest. Similarly, the most distinctive monster cards of \textit{Yu-Gi-Oh!} were only that remarkable because they expressed a deep connection to the main characters of the underlying story (such as the \textit{Dark Magician} card and character \textit{Yugi Muto}, or the \textit{Blue-Eyes White Dragon} card and character \textit{Seto Kaiba}). If everybody would have access to a set of \textit{Blue-Eyes White Dragons}, they would have quickly lost their threat, uniqueness, and narrative identity. However, this is effectively the case every time a metagame slowly converges, as the set of dominant strategies constrains the choice of alternatives too much to still represent identity-centric play (outside casual play). 

This paper is the first in our endeavor to explore a path that allows for both competitive metagame evolution while not compromising player-card relatedness and diversity of play. For the larger part of prior work, PCG approaches had to compromise with limitations in the generation of game mechanics or aesthetics (if not both). With their history of being printed on paper and interpreted by human players and collaborative judgement, TCGs relied heavily on the understanding of natural language to constitute and combine game mechanics, which was difficult to model for traditional approaches of symbolic AI and natural language processing. Yet, with the contemporary rise and promising potential of generative AI, especially with respect to language and image models, this goal is no longer inconceivable. Due to our focus on a player-centric procedural approach, we consider a special type of relatedness that factors in how players relate to the entire process as well as the end result, which we dub \textit{procedural relatedness}.

Hence, this work contributes a dualistic pipeline, an implementation, and an evaluation thereof to the fields of PCG and generative AI in and for games. To assess the feasibility and players' perspectives when encountering such a novel system, we set out to answer the following research questions:
\begin{itemize}
    \item \textbf{RQ1:} How to unite the rapid advancements of generative AI for content generation of multiple facets of design (visual and mechanical)?
    % \item \textbf{RQ2:} How effectively can AI-driven trading card generation realize fitting \textbf{visual aesthetics} with contemporary models?
    \item \textbf{RQ2:} To what degree can such player-centric generation maintain \textbf{representativeness} and \textbf{aesthetics} with respect to the individual's idea?
    \item \textbf{RQ3:} What are suitable iterative strategies to further close the gap between ideation and generation?
\end{itemize}
By answering these questions in the following, we deliver first evidence that \textit{procedural relatedness} in (video) games can be considerably realized through 1) a working technical integration, 2) satisfying visual representation, 3) representative design, and 4) player agency to steer creations towards clear expectation matches.

\section{Related Work}
Only a decade ago, PCG of (trading) card game content was not even a research topic yet, judging from its absence in major review papers of the field. Hendrix et al. aggregated a large corpus of research on PCG within and outside digital games, yet lacking any successful applications on data originating from cards \cite{hendrikx2013procedural}. Similarly, neither Togelius et al. identified any work in this direction concerning symbolic (or search-based) PCG approaches \cite{togelius2011search}, nor did Yannakakis and Togelius acknowledge any PCG approaches for card games that focus on player experience \cite{yannakakis2011experience}. On the commercial side, in 2016, Stone Blade Entertainment released \textit{Solforge}, a card game where cards procedurally evolve during gameplay to create unique strategies \cite{solforge}. In $2018$, designer Richard Garfield introduced \textit{Keyforge}, a card game where deckbuilding is accomplished using PCG steps instead of the player explicitly choosing the set of cards that make up the deck \cite{keyforgeGDC}. This design aims to create a playing environment where all decks are indeed unique -- but far from tailored to the player and/or their individual connection or relation to the content.
In 2018, Summerville et al. review especially ML-Driven PCG at large, and discuss in detail how the generation of TCG cards could be fueled by methods of machine learning. They highlight the crucial need for balance, forms of representing card data structures, and training methods \cite{summerville2018procedural}.
As the most salient and groundbreaking work appears the take of Morgan Milewicz (\textit{RoboRosewater} on \textit{X})\footnote{\url{https://x.com/RoboRosewater}}, %\footnote{\url{https://x.com/RoborosewaterM}}, 
who trained deep recurrent neural networks (RNNs) to learn sequential language data of a \textit{Magic} card corpus, and proceeded to generate and post one card daily \cite{Milewicz2015, Merchant2015}. While he proved that this approach is indeed capable of creating \textit{some} cards, and novelty is almost guaranteed in this probabilistic approach, the generation does not follow a predefined goal or any individual direction, as constrained by the underlying RNN architecture.
Inspired by this success, however, Summerville and Mateas proposed \textit{Mystical Tutor}, in which they utilized Long Short-Term Memory networks (LSTMs) to generate a full card from partial specification as input, overcoming the limitation of not being able to freely generate any part of the card data structure, which opened a path towards individually tailored generation \cite{summerville2016mystical}.
Using the card database of \textit{Hearthstone}, Bhatt et al. simulate different agents with evolving card decks to draw conclusions about dominant strategies and choices \cite{bhatt2018exploring}. In a sense, they communicate a contribution to PCG through the procedurally evaluated deck constructs, yet no generation on the lower level of cards themselves has been made so far.
In contrast, Chen and Guy's approach of \textit{Chaos cards} did indeed generate \textit{Hearthstone}-like cards procedurally, based on grammars 
and automatic evaluation (yet only focusing on logical representation without graphical features)
\cite{chen2020chaos}. Notably, they deployed deep learning to predict the strength of generated cards, which could later be used to ensure balance through proper distribution. 
In order to formalize card game structure and make generation thereof more accessible, several approaches have been taken in the past, such as Font et al.'s card game description language \cite{font2013card}, building on immediate previous advances \cite{thielscher2010general}, or, more recently, van Rozen's unification of the most promising techniques \cite{van2023towards}. While they had PCG as one possible avenue of their work in mind, they mainly targeted traditional (non-trading) card games, in which the main semantics are not defined by local, card-specific language, but symbolic card values and global rule sets.

In other contexts, LLMs have been successfully applied to change \textit{themes} in the setting of a game (see \textit{CrawLLM} \cite{zammit2024crawllm}), which is comparable to our generative image approach for card artwork alternatives that extend existing sets, yet no mechanical knowledge has been incorporated by these authors yet. \textit{LLMaker} goes one step further and offers a framework for prompt-based level design \cite{Gallotta2024llmaker1, gallotta2024consistent}. While this ensures controllability of the output by relying on function calling in the eventual steps, it misses out on the (potentially promising or potentially undesirable) power of emerging mechanics. Games like \textit{1001 Nights} have already incorporated the blend of LLMs and image generation for dynamic creation of contextualized items during play, and lay out open challenges, most notably the compromise between freedom and balance \cite{Sun_Li_Fang_Lee_Asadipour_2023}. 

Maleki and Zhao recently reviewed the intersection of generative AI and PCG \cite{maleki2024procedural}, with most of the approaches still leaning into level generation \cite{schrum2020cppn2gan,kumaran2019generating,kumaran2023end,todd2023level} or free-form language games \cite{zhu2023calypso,nasir2024word2world}.
Very recently, though, first steps towards reasoning and balancing of dynamic (e.g. generated) rules have been investigated, such as in \textit{Baba is You}, where game rules and mechanics are by design chaotically reassembled \cite{van2025baba}. This puts us in an auspicious mood for the applicability and controllability of the relatively uncharted field of LLM-driven mechanics generation. Notably, for the visual generation part, several free or token-based services have established themselves where users can generate Pokémon-like creatures\footnote{\url{https://nokemon.eloie.tech/}} or trading cards\footnote{ \url{https://app.scenario.com/?openAppId=wflow_trading-card-maker}}, yet without any mechanical understanding or generation. To round it up again, no prior work that attempted the generation of cards or similar game content has measured some assessment on relatedness or representativity yet -- which leaves us without a benchmark to compare against, but underlines the novelty of our endeavor, not only in a technical sense, but also with its implications for games user research.

% contextualization
\begin{figure}[!h]
    \centering
    \includegraphics[width=0.8\linewidth]{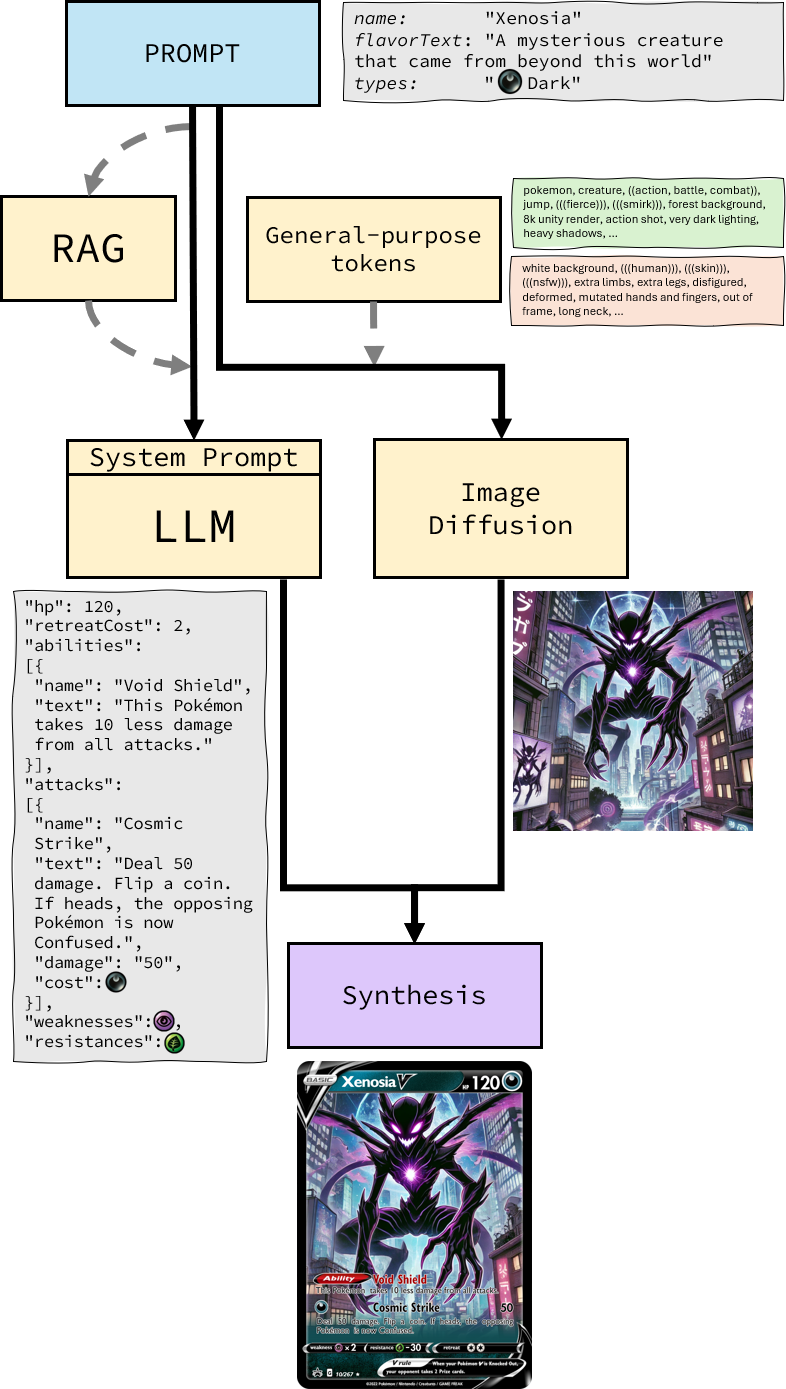}
    \caption{Flowchart of the generation pipeline used in this approach (accompanied by examples). From as little as a name, a type, and a description sentence, a final card is generated (while model and parameter choices are still customizable).}
    \Description{Flowchart of the generation pipeline used in this approach. Starting from the top, a prompt including a name, a type, and a description sentence, is accompanied by examples retrieved using RAG and sent to the LLM. The LLM generates all the necessary card information. The prompt is also sent to the diffusion model along with general-purpose tokens describing the overall style of the image (e.g., pokemon, creature) to generate a visual representation of the card. The two outputs are synthesized to produce the complete card.}
    \label{fig:flowchart}
\end{figure}

\section{Approach}
We tackled the generation of novel cards as a five-step process (cf. Figure \ref{fig:flowchart}): First, we assemble a knowledge base of the mechanical structure of a large variety of cards by querying, filtering, and restructuring public API data. Second, we utilize this ground truth corpus to compute embeddings, which allow for swift alignment of custom prompts to suitable mechanical components in the later retrieval-augmented generation (RAG) step \cite{gao2023retrieval}. Once this is prepared, the following pipeline can employ any compelling LLM for the completion of card data (guided by an adequate system prompt) and any satisfying image model for the parallel artwork generation (supported by customizable auxiliary prompts). In the last step, the aggregated pieces of data are joined and translated into a format that is understandable for one of the readily available Pokémon card creator tools for synthesis, which in turn results in a final print version image of a card.

\begin{figure*}[!h]
    \centering
     \begin{subfigure}{0.24\textwidth}
    \includegraphics[width=\linewidth]{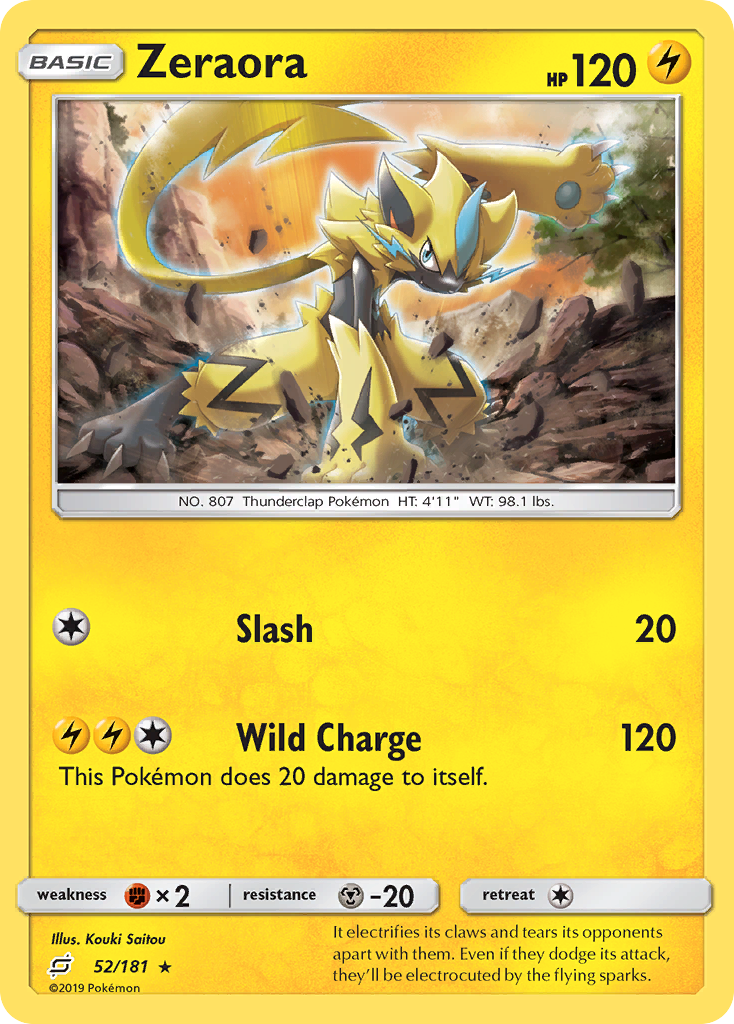}
    \end{subfigure}
\begin{subfigure}{0.74\textwidth}
\begin{lstlisting}[language=json]
{
    "name": "Zeraora",
    "flavorText": "It electrifies its claws and tears its opponents apart with
                   them. Even if they dodge its attack, they'll be 
                   electrocuted by the flying sparks.", 
    "types": ["Lightning"], 
    "hp": "120", 
    "attacks": [{"name": "Slash", "cost": ["Colorless"], "damage": "20"},
                {"name": "Wild Charge", "cost": ["Lightning", "Lightning",
                "Colorless"], "damage": "120", "text": "This Pokemon does 
                20 damage to itself."}], 
    "weaknesses": [{"type": "Fighting", "value": "x2"}], 
    "resistances": [{"type": "Metal", "value": "-20"}], 
    "convertedRetreatCost": 1, 
    "artist": "Kouki Saitou", 
    "images": {"large": "https://images.pokemontcg.io/sm9/52_hires.png"}
}
\end{lstlisting}
    % "id": "sm9-52", 
    % "supertype": "Pok\u00e9mon", 
    % "subtypes": ["Basic"], 
    % "retreatCost": ["Colorless"], 
    % "number": "52", 
    % "rarity": "Rare Holo", 
    % "nationalPokedexNumbers": [807], 
    \end{subfigure}
    
    \caption{Scan of an original Pokémon card, and the according data structure retrieved from the database (some fields excluded for brevity). Using 993 structures, we computed embeddings that are majorly defining the shape of the final LLM output.}
    \Description{Scan of an original Pokémon card, and the according data structure as retrieved from the database in a JSON format. Primary fields are \textit{name}, \textit{flavorText}, \textit{types}, \textit{hp}, \textit{attacks}, \textit{weaknesses}, \textit{resistances}, \textit{convertedRetreatCost}, \textit{artist}, and \textit{images}}
    \label{fig:zeraora}
\end{figure*}

\subsection{Database Mining}
To have a dataset that is as rich as possible to start with, we queried the publicly available API of the \textit{Pokémon TCG Developer Portal} \footnote{\url{https://dev.pokemontcg.io/}} for structured information about all Pokémon trading cards published at the time of retrieval, which resulted in $15,411$ distinct cards (e.g., see Figure \ref{fig:zeraora}). Many of them referred to reprinted versions (in newer sets), alternative forms, event editions, or other forms of duplicates, which we carefully excluded to not overfit on older or more popular Pokémon species -- hence, we focused on basic Pokémon card versions until the $9^{th}$ generation for this first evaluation of our approach. Data for the resulting $993$ unique cards was downloaded and split into separate JSON files and semantic chunks thereof to enable contextualization through embeddings towards RAG (see Section \ref{sec:Embeddings}). To improve their suitability for the task ahead, we ruled out data fields that were irrelevant for the mechanical description of a card and sorted the order of the remaining fields to ease the eventual completion task of the LLM (see Section \ref{sec:LLM}). The processed, embedding-ready dataset is available in the open-source repository accompanying this work %\footnote{ \url{https://github.com/JohannesPfau/generativePokemonTCG/tree/main/pokemonData}}, 
or can be updated/reproduced following the accompanying data retrieval script%\footnote{[redacted for review]}.
\footnote{\url{https://github.com/JohannesPfau/generativePokemonTCG/blob/main/pokemonDBtoJson.py}}.

\subsection{RAG}
\label{sec:Embeddings}
As outlined in the initial part of the section, we utilize custom embeddings on the basis of the collected dataset. This allows us to quickly contextualize data from an input prompt within similar examples of the knowledge base (by localizing a subset of cards for which the distance to the input is smallest in the embedding space). For the seamless integration during the evaluation of this approach, we used \textit{LangChain} embeddings \cite{folstad2019chatbots} (to be more precise: the \textit{nomic-embed-text-v1.5} model), but leave the choice of the particular implementation open. In the proposed pipeline, we implement RAG by calculating the cosine similarity of the input prompt (consisting of \textit{name}, \textit{flavorText} and \textit{types}) against the JSON structures of the existing Pokémon cards (e.g., see Figure \ref{fig:zeraora}), and the closest matches are included in the succeeding LLM prompt to supply it with exemplary properties and mechanics that fit similar already published Pokémon cards.

\subsection{LLM}
\label{sec:LLM}
\begin{figure}[!h]
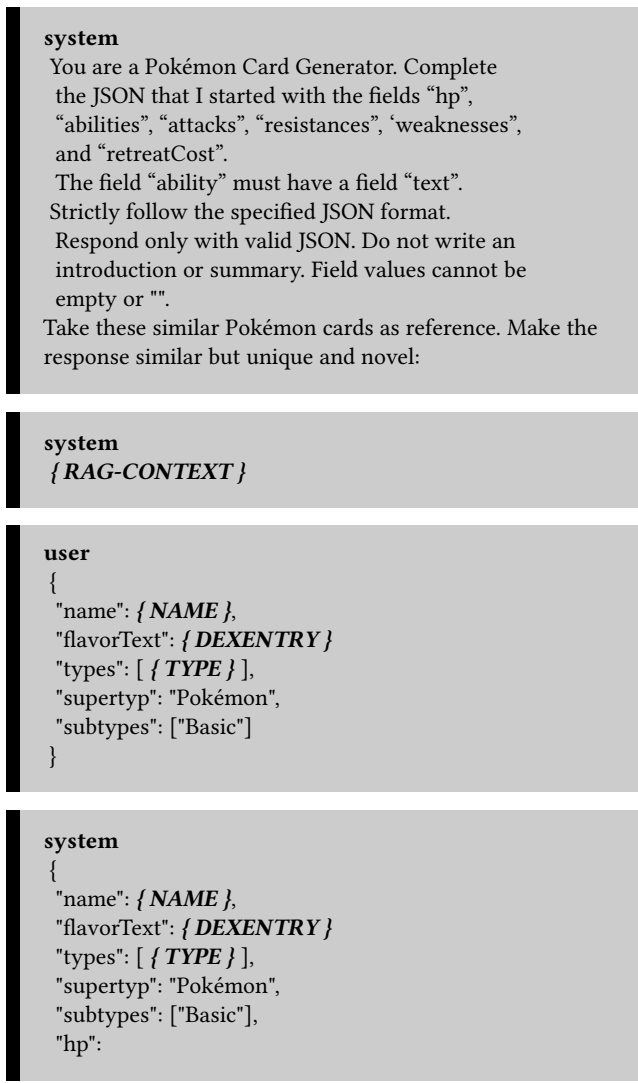

\begin{leftbar} 
%$<$$|$begin\_of\_text$|$$>$
%~~$<$$|$start\_header\_id$|$$>$\textbf{system}$<$$|$end\_header\_id$|$$>$
\textbf{system} \\
~~~~You are a Pokémon Card Generator. Complete \\
\phantom{~~~~} the JSON that I started with the fields ``hp'',\\ \phantom{~~~~} ``abilities'', ``attacks'', ``resistances'', `weaknesses'',\\
\phantom{~~~~} and ``retreatCost''.\\
\phantom{~~~~} The field ``ability'' must have a field ``text''.\\
%\phantom{~~~~} $<$$|$eot\_id$|$$>$
%\\
%
%~~$<$$|$start\_header\_id$|$$>$\textbf{user}$<$$|$end\_header\_id$|$$>$
%
~~~~Strictly follow the specified JSON format.\\
\phantom{~~~~} Respond only with valid JSON. Do not write an\\
\phantom{~~~~} introduction or summary. Field values cannot be \\
\phantom{~~~~} empty or "".
%$<$$|$eot\_id$|$$>$
\\
Take these similar Pokémon cards as reference. Make the response similar but unique and novel:
%~~$<$$|$start\_header\_id$|$$>$\textbf{assistant}$<$$|$end\_header\_id$|$$>$
%\{
\end{leftbar}

\begin{leftbar} 
\textbf{system} \\
~~~~\textbf{\textit{\{ RAG-CONTEXT \}}}
\end{leftbar}

\begin{leftbar} 
\textbf{user} \\
~~~~\{ \\
\phantom{~~~~} "name": \textbf{\textit{\{ NAME \}}},\\
\phantom{~~~~} "flavorText": \textbf{\textit{\{ DEXENTRY \}}} \\
\phantom{~~~~} "types": [\textbf{\textit{ \{ TYPE \}}} ],\\
\phantom{~~~~} "supertyp": "Pokémon",\\
\phantom{~~~~} "subtypes": ["Basic"]\\
~~~~\}
\end{leftbar}

\begin{leftbar} 
\textbf{system} \\
~~~~\{ \\
\phantom{~~~~} "name": \textbf{\textit{\{ NAME \}}},\\
\phantom{~~~~} "flavorText": \textbf{\textit{\{ DEXENTRY \}}} \\
\phantom{~~~~} "types": [\textbf{\textit{ \{ TYPE \}}} ],\\
\phantom{~~~~} "supertyp": "Pokémon",\\
\phantom{~~~~} "subtypes": ["Basic"],\\
\phantom{~~~~} "hp": 
\end{leftbar}
    \caption{System prompt used for the structured card description generation in this work.
    \textit{system} initializes the main purpose of the instruct model in the session and dynamically includes the semantically closest cards of the knowledge base as RAG context. \textit{user} formalizes the prompt parameters into the desired structure, which is once again repeated by \textit{system} with its JSON structure deliberately left open-ended to trigger completion of the missing fields.}
    \label{fig:systemPrompt}
\end{figure}

While the choice of the particular language model is as well debatable (and only an artifact of our time), we chose to deploy %\textit{Llama-3-8B-Instruct} \cite{llama3modelcard}
\textit{Qwen3-14B} \cite{qwen3technicalreport} 
for its convenient trade-off between established code understanding, thinking/reasoning capabilities, output standardization (in the form of a goal JSON schema), suitable context window, and the objective to be run on consumer devices, which furthermore yielded convincing results on initial test runs. In general, the role of the LLM in this pipeline is to complete an underspecified JSON file in the format of the formerly established database, where \textit{name}, \textit{flavorText} and \textit{types} are prompted by the user. Most importantly, the health points of the Pokémon (\textit{hp}), its \textit{abilities}, \textit{attacks}, \textit{resistances}, and \textit{weaknesses} should be predicted, as indicated in the specified system prompt (see Figure \ref{fig:systemPrompt}).

To be more precise, a Pokémon's passive \textit{abilities} field is a compound object as well, which is defined by its \textit{name} and effect description (\textit{text}). This is similar to their active \textit{attacks}, which, on top of that, add the Energy \textit{cost} required to activate, as well as their \textit{damage} output. Since the resulting dictionary does not always contain the input information again, this data is joined, cleaned, and verified for LLM-induced issues such as incomplete information or invalid structure%\footnote{[redacted for review]}.
\footnote{\url{https://github.com/JohannesPfau/generativePokemonTCG/blob/main/pokemonGenerate.py}}. 

\subsection{Image Diffusion}
In parallel to the information completion through the LLM, a generative image model is prompted with the same input fields as before, only enriched by a series of general-purpose tokens%\footnote{[redacted for review]}. 
\footnote{\url{https://github.com/JohannesPfau/generativePokemonTCG/blob/main/pokemonGenerateImage.py}}.
These additional (positive and negative) prompts were composed by taking inspiration from established tokens that worked well with generating Pokémon images in the community around the selected model. For this iteration, we ran a local installation of \textit{ComfyUI} %\footnote{https://github.com/comfyanonymous/ComfyUI} 
and deployed \textit{FLUX.1-dev-Q8} as a diffusion model that served to be suitable for the task for its high quality and speed, but emphasize that this can be easily replaced with potentially more sophisticated models and/or fine-tunings for future use. To fine-tune the diffusion process to match the aesthetic of Pokémon art, we used a well-balanced combination of the \textit{Niji} and the \textit{Pokémon: Ken Sugimori style} low-rank adaptation model (LoRA) that adequately suited the style \footnote{https://civitai.com/models/671563 ; https://civitai.com/models/871757}. The final ComfyUI workflow is enclosed in our open-source repository for reproducibility.%\footnote{[redacted for review]}.

\subsection{Synthesis}
In the final step, the unified JSON (initial prompt, LLM response) is combined with the generated image and translated into a different JSON representation, now suitable to function as input for the print version card generator. In this case study, we relied on the web-based \textit{Pokécardmaker}\footnote{\url{https://pokecardmaker.net/creator}} as it presented a readily available option capable of importing JSON information, but admit that potential local options (such as the established \textit{Magic Set Editor}\footnote{\url{https://magicseteditor.boards.net/}}) could keep this approach completely offline.

\section{Evaluation}
To capture a broader creative range (than just that of the authors), to measure the \textit{quality} of the generated mechanics and aesthetics, to investigate if the desired mechanics and images were \textit{representative} with respect to the individual ideation, and to reduce bias in qualitatively interpreting the feasibility of the approach, we formulated a mixed-methods evaluation design and reached out to a target group that we deemed appropriate to sufficiently understand both the pipeline as well as the validity regarding the Pokémon TCG and its mechanics. 

\subsection{Participants}
We recruited ($n=49$) Master's students ($75.5\%$ male, $24.5\%$ female) visiting a topically related course at the local university of the authors, who were primarily enrolled in \textit{Game and Media Technology} and \textit{Artificial Intelligence} Master's programs. %The generally above-average technical literacy of our participants facilitated the evaluation process and removed the need for the --unnecessary for this study-- development of a user interface tailored to the general public.

\subsection{Procedure}
We presented our approach (as published in the referenced repository) and asked them to generate four different cards each. To realize that, they could choose to install the pipeline on their local machines or use a provided, readily available server (which responded in $\sim20$ seconds generation time, computed mainly on an NVIDIA RTX 5090). They could freely configure and arrange their input prompts, and were given the option to customize the image and language models, parameters, and the system prompt. Eventually, they submitted $196$ cards (e.g., Figures \ref{fig:teaser}, \ref{fig:flowchart}, and \ref{fig:emergentVsFlawed}), which are fully accessible in the supplementary repository. 
%\footnote{\url{https://github.com/JohannesPfau/generativePokemonTCG/tree/main/results}}. 
% In the following, we report on our lessons learned and summarize trends of promising outcomes as well as systematic shortcomings of this approach.
As soon as they were satisfied with their generated card(s) -- or if they gave up because of unsatisfactory results -- they were asked to complete a questionnaire for all of their generations, as outlined in the following section.
After the study concluded, quantitative metrics were aggregated and interpreted with respect to their scales, whereas for qualitative statements, two authors first independently applied thematic analysis with open codes \cite{braun2006using} and reviewed their codes subsequently to make final agreements.

\begin{figure*}[h!]
    \centering
    \includegraphics[width=0.141\linewidth]{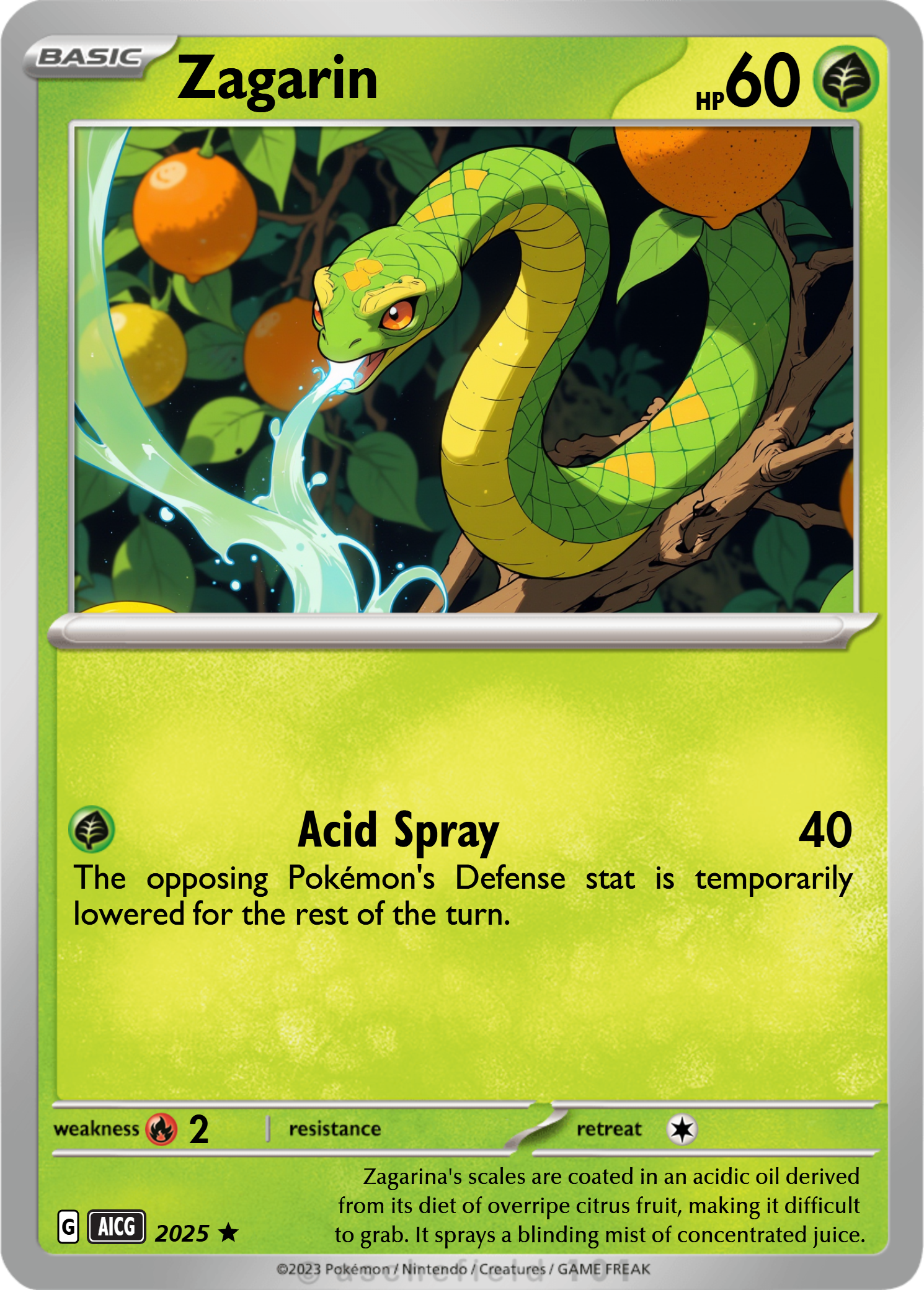}\includegraphics[width=0.141\linewidth]{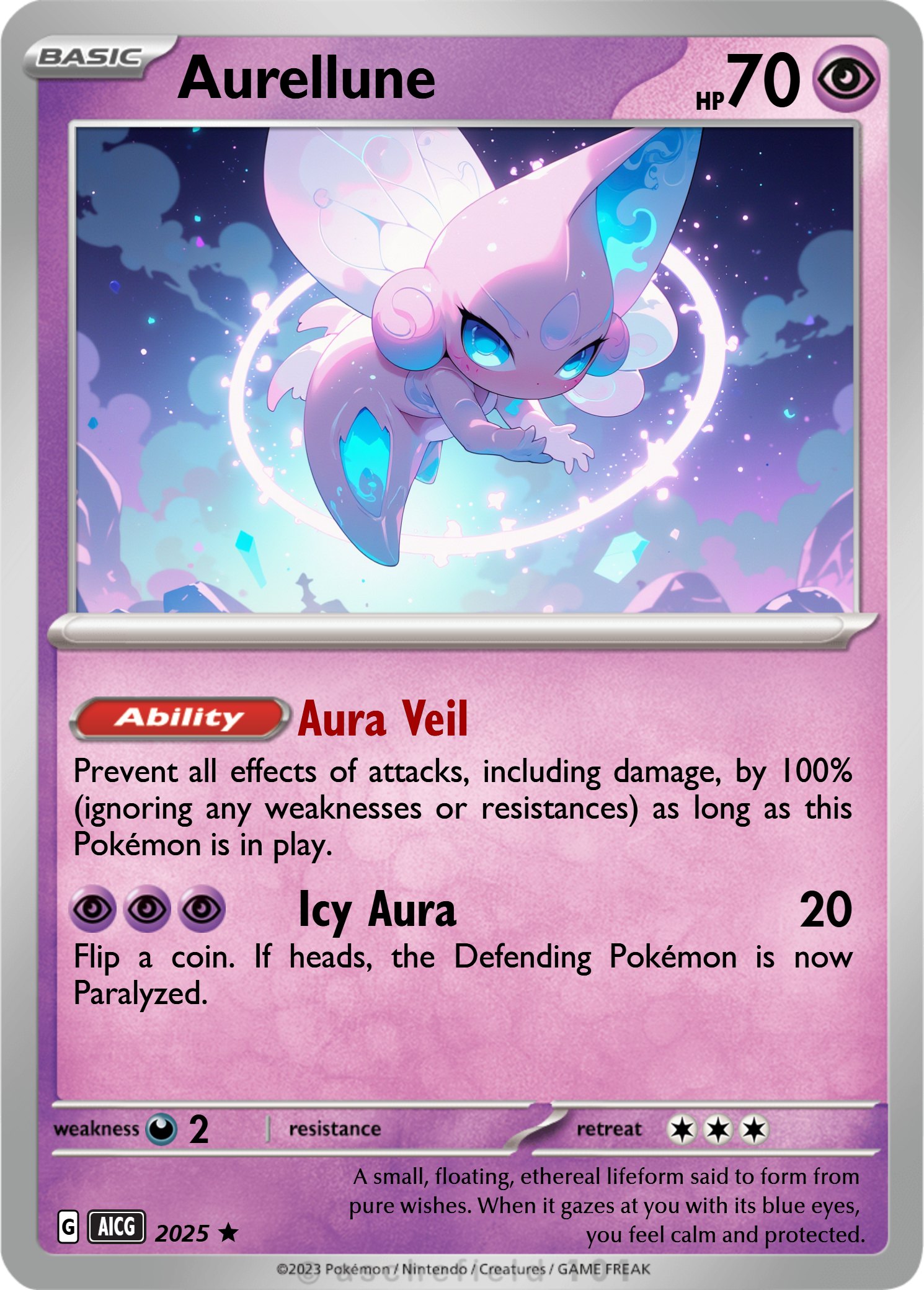}\includegraphics[width=0.141\linewidth]{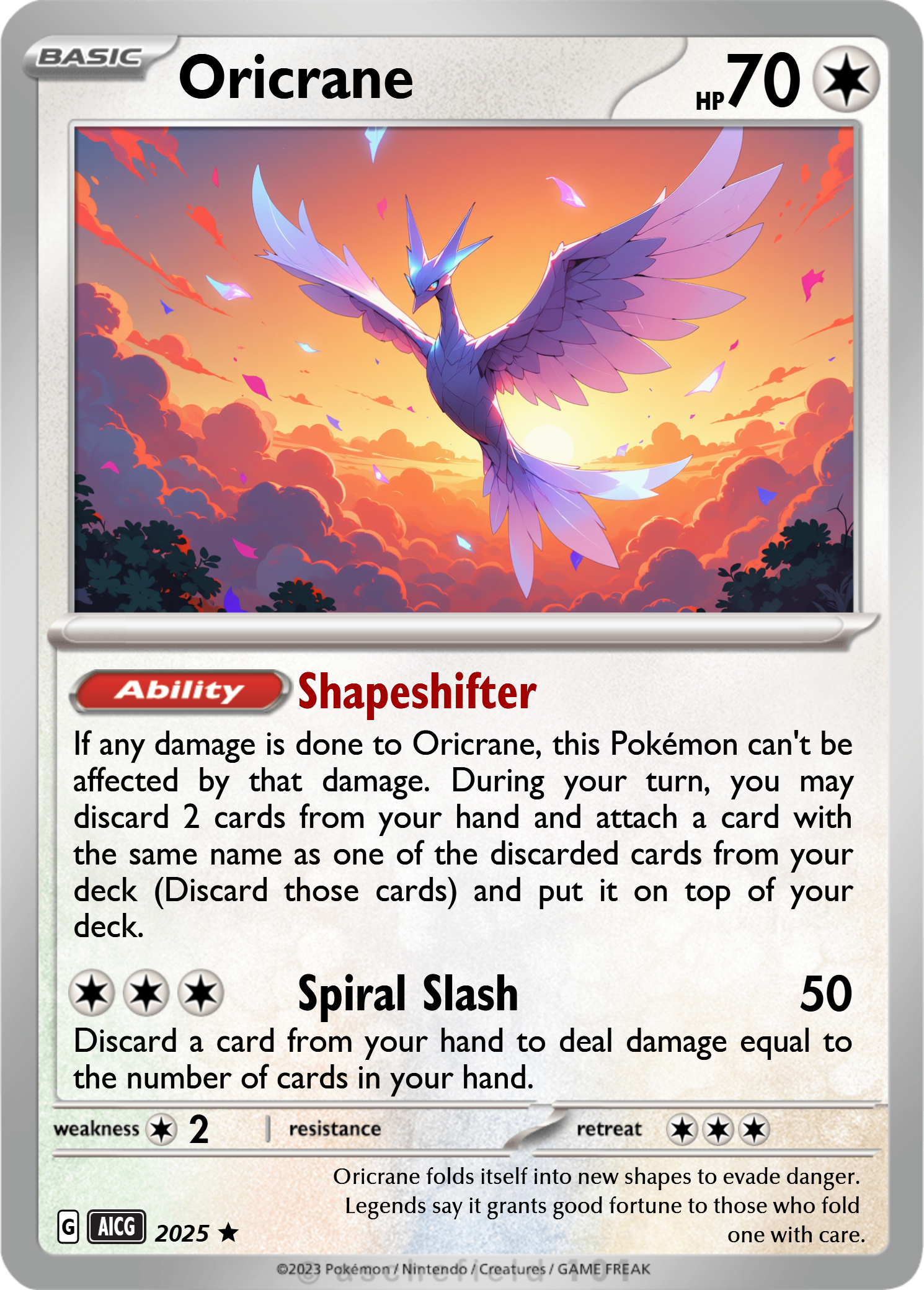}\includegraphics[width=0.141\linewidth]{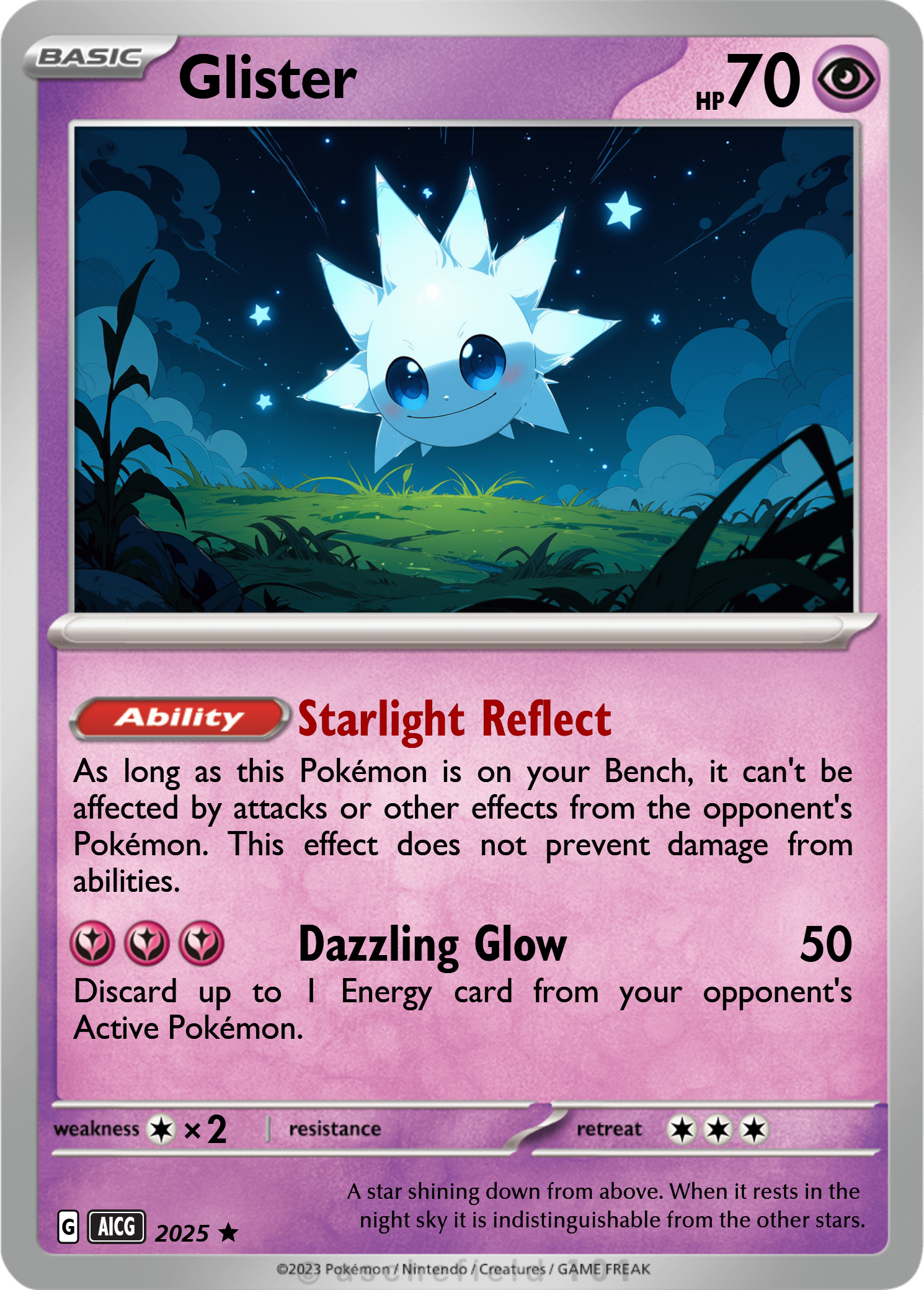}\includegraphics[width=0.141\linewidth]{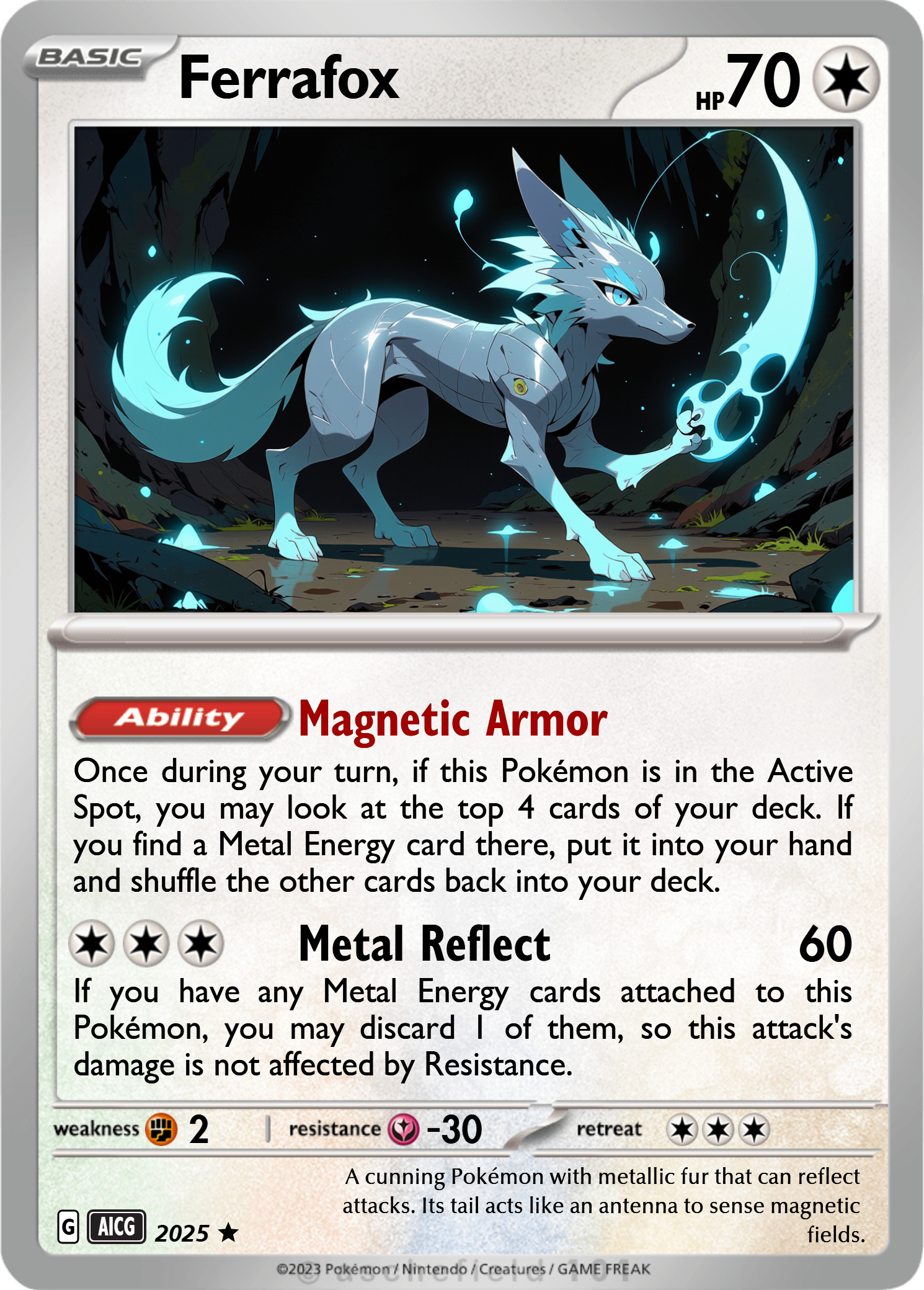}\includegraphics[width=0.141\linewidth]{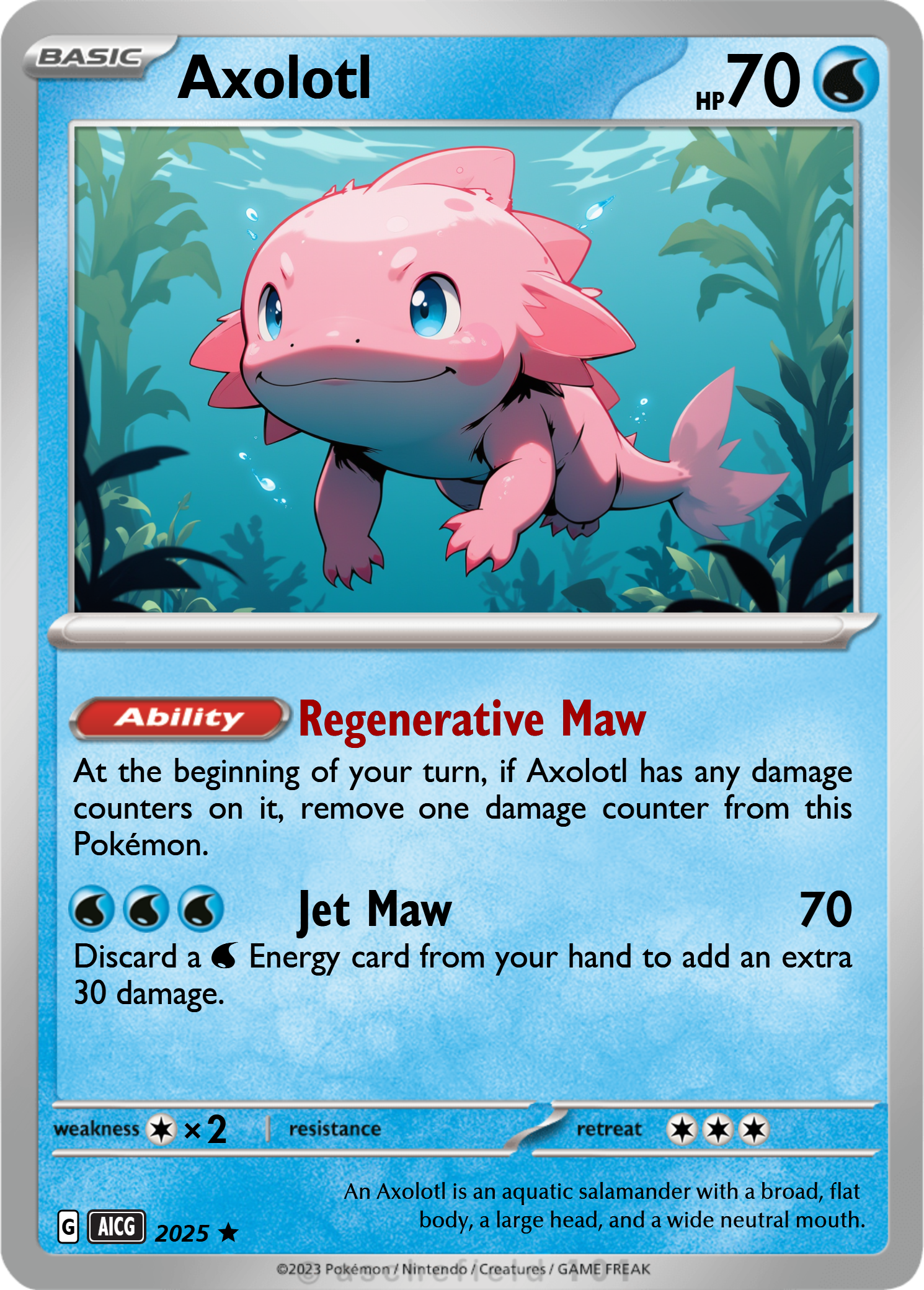}\includegraphics[width=0.141\linewidth]{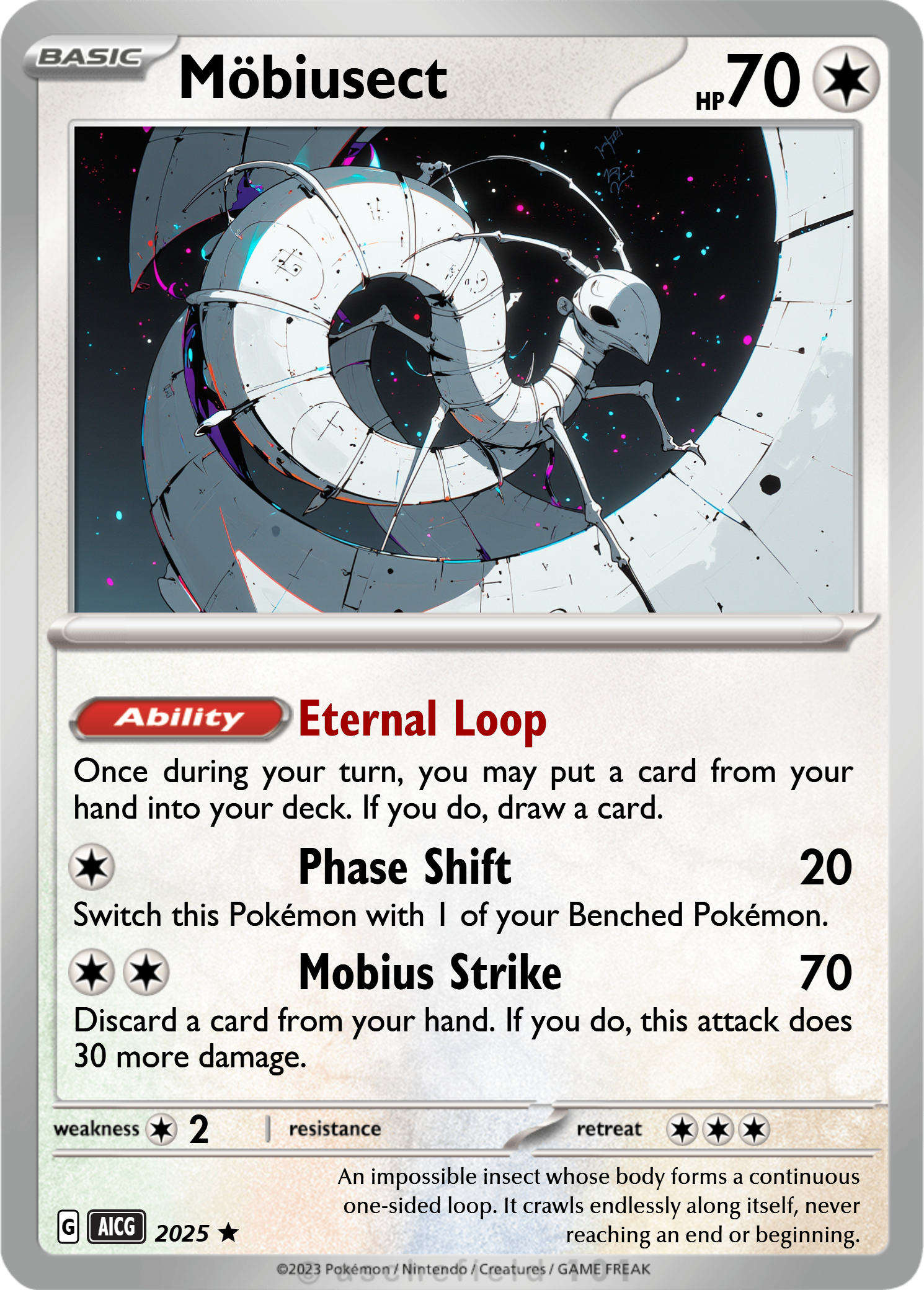}
    \caption{Examples of generations that show common flaws (under-specification in \textit{Zagarin}; imbalance in \textit{Aurellune}; idiosyncratic repetition in Oricrane), but also \textit{emergent game mechanics} (Glister: from being \textit{indistinguishable in night sky} in its description $\rightarrow$ a protective ability during bench placement; Ferrafox: from being prompted to be metallic $\rightarrow$ attracting and repelling mechanics are formed in ability and attack; Axolotl: from only the inherent world knowledge that axolotls have remarkable regeneration properties $\rightarrow$ a mechanically correct regeneration ability emerges; Möbiusect: from an abstract design of repetition, loops and iteration $\rightarrow$ the model fittingly generates mechanics to loop through the deck, swap through active Pokémon, and connects drawing/discarding hand card mechanics).}
    \Description{Examples of generations that show common flaws (under-specification in \textit{Zagarin}; imbalance in \textit{Aurellune}; idiosyncratic repetition in Oricrane), but also \textit{emergent game mechanics} (Glister: from being \textit{indistinguishable in night sky} in its description $\rightarrow$ a protective ability during bench placement; Ferrafox: from being prompted to be metallic $\rightarrow$ attracting and repelling mechanics are formed in ability and attack; Axolotl: from only the inherent world knowledge that axolotls have remarkable regeneration properties $\rightarrow$ a mechanically correct regeneration ability emerges; Möbiusect: from an abstract design of repetition, loops and iteration $\rightarrow$ the model fittingly generates mechanics to loop through the deck, swap through active Pokémon, and connects drawing/discarding hand card mechanics).}
    \label{fig:emergentVsFlawed}
\end{figure*}

\subsection{Measures}
To capture \textit{Aesthetics} and \textit{Representativeness} of card images and mechanics, we posed the following custom 5-point Likert questions for every single card created:

\begin{itemize}
    \item \textbf{Aesthetics}: \\
    How satisfied were you with the \textbf{visual aesthetic} of the (final) outcome for your card?\\
    {\small \textit{Very Dissatisfied / Dissatisfied / Neutral / Satisfied / Very Satisfied}}
    \item \textbf{Representativeness (Image)}: \\
    How closely did the (final) outcome \textbf{image} resemble what you had in mind for the card? \\
    {\small \textit{Very Poor / Poor / Neutral / Good / Very Good}}
    \item \textbf{Representativeness (Mechanics)}: \\How much did the (final) outcome of the \textbf{attacks/ability/rest of card values} suit the concept of the card?\\
    {\small \textit{Very Poor / Poor / Neutral / Good / Very Good}}
\end{itemize}

On top of these quantitative scores, we added open-ended qualitative questions, which would subsequently be coded and interpreted:
\begin{itemize}
    \item Briefly describe in your own words what you "expected" to get from your card (\textbf{ideation}), 
    \item what the approach generated in the FIRST try (\textbf{creation}), 
    \item and how you potentially \textbf{adapted} the prompt, script, or your idea.
    \item For the final outcomes, would you say your generated card follows \textbf{your own idea} - or did your idea/concept change based on what you generated?
    \item Additional Remarks
\end{itemize}

Lastly, the authors went over all $196$ generated outcomes and classified trends of systematic strengths and flaws of the approach.

% \begin{figure*}[!h]
%     \centering
%     \includegraphics[width=0.168\linewidth]{img/fail/Mossprout_Card.png}\includegraphics[width=0.168\linewidth]{img/fail/undertaker_card.png}\includegraphics[width=0.168\linewidth]{img/fail/Tidefin.png}\includegraphics[width=0.168\linewidth]{img/fail/Vicar.png}\includegraphics[width=0.168\linewidth]{img/fail/Pizzaton.png}\includegraphics[width=0.168\linewidth]{img/fail/Ninju.png}
%     \caption{Sample outcomes of generated cards with systematic flaws, as highlighted in the discussion.}
%     \label{fig:cardsFlawed}
% \end{figure*}

\section{Results}
\begin{figure}[h!]
    \centering
    \includegraphics[width=0.6\linewidth]{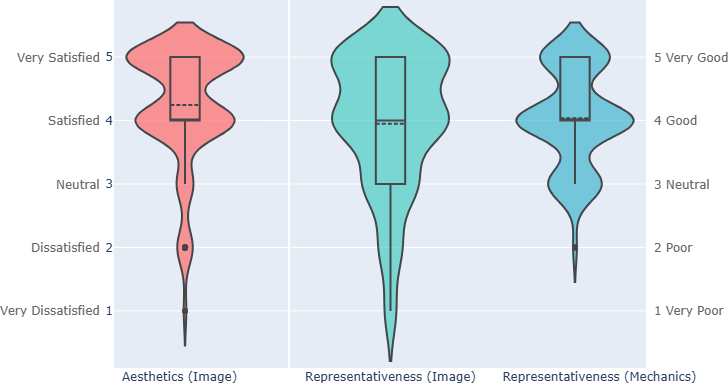}
    \caption{Violin Chart of the quantitative metrics (Aesthetics and Representativeness) assessing $196$ cards of our pipeline.}
    \Description{Violin Chart of the quantitative metrics (Aesthetics and Representativeness) assessing $196$ cards of our pipeline. High satisfaction on image aesthetics and representativeness of mechanics is indicated by the majority of responses falling between $4$ (i.e., satisfied and good, respectively) and $5$ (i.e., very satisfied and very good, respectively). For visual representativeness, scores are high, between $3$ (i.e., good) and $5$ (i.e., very good).}
    \label{fig:quantResults}
\end{figure}

In terms of quantitative metrics, the set of $196$ generated cards was rated as \textit{Satisfying} ($M=4.25, SD=0.9$) in terms of visual aesthetics, as \textit{Good} ($M=3.95, SD=1.07$) in terms of visual representativeness, and as \textit{Good} ($M=4.04, SD=0.75$) in terms of mechanical representativeness (see Figure \ref{fig:quantResults}).

\begin{figure}[h!]
    \centering
    \includegraphics[width=0.6\linewidth]{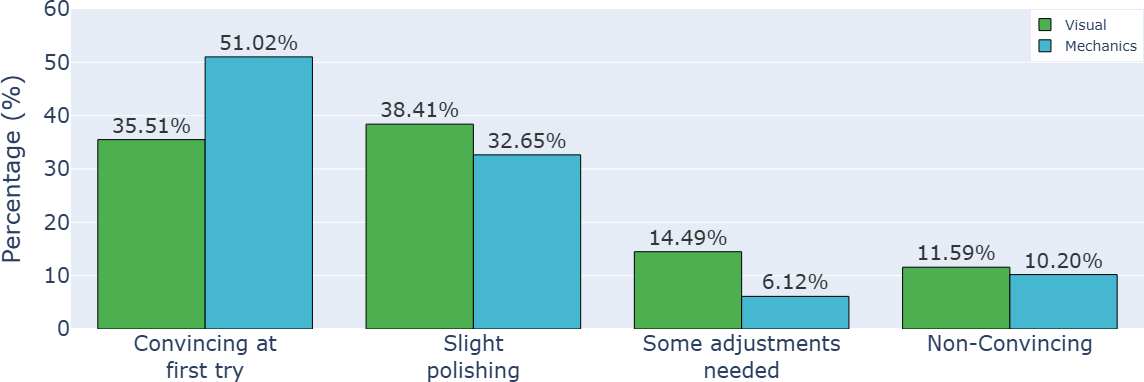}
    \caption{Distribution of qualitative codes for the \textit{Expectation Match} theme with regards to how well the generation matched the participants' visual (green) and mechanical (blue) expectations.}
    \Description{Distribution of qualitative codes for the \textit{Expectation Match} theme with regards to how well the generation matched the participants' visual and mechanical expectations. $35.51\%$ felt convinced at first try about the visuals of the generation and $51.02\%$ felt similarly about the mechanics. $38.41\%$ felt that slight polishing was required for the visuals and $32.65\%$ for the mechanics. $14.49\%$ required some adjustments to the visuals and $6.12\%$ to the mechanics. $11.59\%$ felt that the final result was non-convincing visually and $10.20\%$ felt the same about the mechanics.}
    \label{fig:qual_expectationMatch}
\end{figure}

\begin{figure}[h!]
    \centering
    \includegraphics[width=0.6\linewidth]{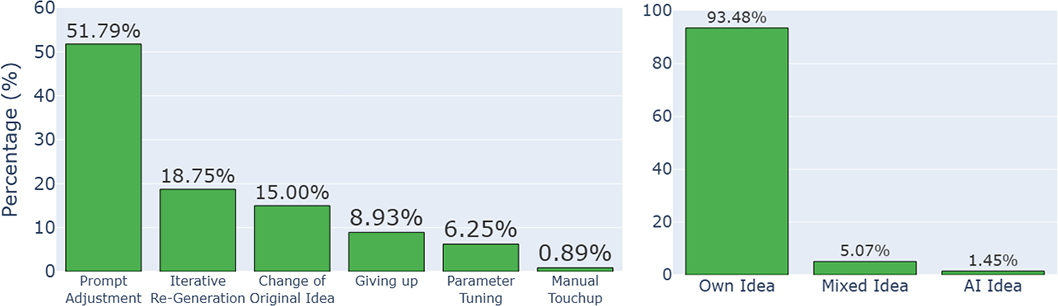}
    \caption{Distribution of qualitative codes for the \textit{Adaptation Method} (left) and the \textit{Idea Attribution} (right) themes.}
    \Description{Distribution of qualitative codes for the \textit{Adaptation Method} and the \textit{Idea Attribution} themes. $51.79\%$ used prompt adjustment, $18.75\%$ used iterative re-generation, $15.00\%$ changed the original idea, $8.93\%$ gave up, $6.25\%$ tuned some of the parameters, and $0.89\%$ manually touched up the result. $93.48\%$ reported that the generation was their own idea, $5.07\%$ a mixed idea, and $1.45\%$ an AI idea.}
    \label{fig:qual_adaptation_ideaSource}
\end{figure}

Regarding qualitative analysis, the final themes and codes that emerged from the participant's statements were compiled in Figure \ref{fig:qualitativeResults}. For a comparative overview, Figure \ref{fig:qual_expectationMatch} contrasts the code distributions for \textit{Expectation Match} (visual as well as mechanical), and Figure \ref{fig:qual_adaptation_ideaSource} contrasts first the distribution of the different \textit{Adaptation Methods} and then the distribution of the \textit{Idea Attribution} responses.

\begin{itemize}
    \item \textbf{Expectation Match (visual):}\\
    {\small \textit{Whether the generated artwork matched the participant's visual expectation}}
    \begin{itemize}
        \item (35.5\%) \textbf{Convincing at first try:} If the card artwork was compelling right away
        \item (38.4\%) \textbf{Slight polishing:} If the card artwork was acceptable right away, but improved in slight iteration
        \item (14.5\%) \textbf{Some adjustment needed:} If the card artwork was not acceptable right away, but the adaptation method fixed this
        \item (11.6\%) \textbf{Non-Convincing:} If the participant did not end up with satisfying artwork, even after iteration
    \end{itemize}
    
    \item \textbf{Expectation Match (mechanical):}\\
    {\small \textit{Whether the generated functional design matched the participant's conceptual expectation}}
    \begin{itemize}
        \item (51.0\%) \textbf{Convincing at first try:} If the card's mechanical design was compelling right away
        \item (32.7\%) \textbf{Slight polishing:} If the card's mechanical design was acceptable right away, but improved in slight iteration
        \item ~~(6.1\%) \textbf{Some adjustment needed:} If the card's mechanical design was not acceptable right away, but the adaptation method fixed this
        \item (10.2\%) \textbf{Non-Convincing:} If the participant did not end up with satisfying mechanics, even after iteration
    \end{itemize}
    
    \item \textbf{Adaptation Method (if necessary):}\\
    {\small \textit{The participant's strategy to amend (potential) mismatches between ideation and generation}}
    \begin{itemize}
        \item (51.8\%) \textbf{Prompt Adjustment:} Changing the flavor text, Pokédex entry, or part of the system prompt
        \item (18.8\%) \textbf{Iterative Re-Generation:} Simply trying again, only changing the initial random seed
        \item (13.4\%) \textbf{Change of Original Idea:} Deviating from the ideation in favor of a direction from the generator
        \item ~~(8.9\%) \textbf{Giving Up:} Inability to arrive at a generation that matches the expectations of the idea
        \item ~~(6.3\%) \textbf{Parameter Tuning:} Changing numerical pipeline variables to, e.g., constrain/extend creative output
        \item ~~(0.9\%) \textbf{Manual Touchup:} Adjusting the image or mechanics with external tools or editors
    \end{itemize}
    
    \item \textbf{Idea Attribution:}\\
    {\small \textit{Whether the overall concept of the generation followed the creator, the model, or both}}
    \begin{itemize}
        \item (93.5\%) \textbf{Own Idea:} If the card design resembles the original intention of the creator
        \item ~~(1.4\%) \textbf{AI Idea:} If the card design was mainly driven by the model
        \item ~~(5.1\%) \textbf{Mixed Idea:} If the participant let the model actively extend/complement vague own ideas
    \end{itemize}
    
    \item \textbf{Approach Appreciation:}\\
    {\small \textit{In case the participant noted to like/dislike our approach in general}}
    \begin{itemize}
        \item (11x) \textbf{Positive:} If they explicitly stated to like it
        \item ~~~(0x) \textbf{Negative:} If they explicitly stated to dislike it
    \end{itemize}
    \item \textbf{Miscellaneous Statement Themes:}
    \begin{itemize}
    \item (14x) \textbf{Expectations Exceeded:} If the visual result was not only convincing, but went above and beyond
    \item ~~(7x) \textbf{Preferred Generated to Initial Expectation:} If the generated design deviated from the original idea, but improved it only
    \item (10x) \textbf{Model Limitations:} If the participant blamed the model for failing to have a good understanding of their idea (in contrast to condemning the whole approach/pipeline)
    \item ~~(7x) \textbf{Pokémon Unfamiliarity:} If the participant felt not qualified to judge the mechanical design
    \end{itemize}
    \begin{figure}[h!]
    \caption{Themes and (distribution of) codes from the qualitative analysis. Distinct codes that correspond to a theme are reported relatively to other code distributions within the theme \textbf{(\%)}, whereas miscellaneous categories are reported in absolute counts \textbf{(x)}.}
    \Description{}
    \label{fig:qualitativeResults}
\end{figure}
\end{itemize}

Regarding the post-hoc classification of systematic flaws of the mechanical outcomes, Figure \ref{fig:emergentVsFlawed} exemplifies the most commonly appearing trends among the non-convincing results: \textit{under-specified} mechanics (that need further parameterization in order to be interpretable; \textit{imbalanced} that deviate drastically from common norms (both in the direction of rendering the card unviable as well as too overpowered); and several kinds of \textit{hallucinations} (such as unnecessary repetition, reference to non-existing game mechanics, or conceptual clashes with the original design).
Even so, empty responses, missing fields, or syntax errors did not occur across all output of the pipeline, thanks to the schema-centric output standardization of the model.

On the other hand, Figure \ref{fig:emergentVsFlawed} simultaneously also depicts examples of \textit{emergent mechanics}, so successful translations from natural language concepts or semantics into playable mechanics -- either by mapping the input description to the output structure (e.g., \textit{Glister}) or by enriching the concept of the idea with implicit world knowledge embedded in the foundational model (e.g., \textit{Axolotl}).

%were carefully disassembled towards the themes of a) \textit{if the outcome matches the participant's expectation} (seperately for visuals and mechanics), b) \textit{what adaptation method they used in case of mismatches}, c) \textit{whether the final outcome resembled their idea, or rather the understanding of the model, or a blend in-between}, d) \textit{general opinion of the approach}

\section{Discussion}
In the following, we return to our initially posed research questions, and aim to answer them based on the just listed results.

% \subsubsection*{\textbf{RQ2:}  How effectively can AI-driven trading card generation realize fitting \textbf{visual aesthetics} with contemporary models?}~\\
% As our pipeline is comprised of several style-guiding principles (general-purpose tokens, LoRAs, and a system prompt tailored to the problem domain), the resulting set of all $196$ cards followed a very consistent visual theme (e.g., see Figures \ref{fig:teaser} and \ref{fig:emergentVsFlawed}). This led to a large degree of visual satisfaction among the participants ($M=4.25$ out of $5$), from which the majority of results were convincing right away, in the first few tries, or could be steered into the desired outcome successfully ($88.4\%$ total). 
% In fact, almost $30\%$ of our participants expressed that the visual outcomes even \textit{exceeded their expectations} (without being explicitly asked about this), even though they already had a high literacy of generative AI and game design.
% Only $11.6\%$ of card generation attempts were rated to be visually non-convincing, which suggests a high feasibility of the approach at first shot -- however, seeing every $10th$ artwork fail certainly renders this approach as not entirely production-ready \textit{yet}. With rapidly improving diffusion (as well as multi-modal) models and evolving our approach together with a larger public audience, we are positive about further polishing this first promising implementation.

\subsubsection*{\textbf{RQ2:}  To what degree can such player-centric generation maintain \textbf{representativeness} and \textbf{aesthetics} with respect to the individual's idea?}~\\
Regarding visual \textbf{aesthetics}, our pipeline joined several style-guiding principles (general-purpose tokens, LoRAs, and a system prompt tailored to the problem domain), which resulted in a set of $196$ cards that followed a very consistent visual theme (e.g., see Figures \ref{fig:teaser} and \ref{fig:emergentVsFlawed}). This led to a large degree of visual satisfaction among the participants ($M=4.25$ out of $5$), from which the majority of results were convincing right away, in the first few tries, or could be steered into the desired outcome successfully ($88.4\%$ total). 
In fact, almost $30\%$ of our participants expressed that the visual outcomes even \textit{exceeded their expectations} (without being explicitly asked about this), even though they already had a high literacy of generative AI and game design.
Only $11.6\%$ of card generation attempts were rated to be visually non-convincing, which suggests a high feasibility of the approach at first shot -- however, seeing every $10th$ artwork fail certainly renders this approach as not entirely production-ready \textit{yet}. With rapidly improving diffusion (as well as multimodal) models and evolving our approach together with a larger public audience, we are positive about further polishing this first promising implementation.

While the former point concerned the general visual quality for the application case, we were specifically interested in \textbf{representativeness}, or how close the generated outcome deviated from the idea in the creator's mind, as one potentially central aspect of the initially posed \textit{procedural relatedness} to one's own playable content. Translating the image inside a user's mind onto screen (or paper) is a non-trivial task for generative AI (and art as a whole), and has been the subject of many studies \cite{nayak2025culturalframes, petsiuk2022human, bakr2023hrs}. For the mechanical part, to the best of our knowledge, no prior study assessed the creator's gap between ideation and generation. Hence, akin to previous work that tackled representativeness or alignment in image generation \cite{ma2024subject}, we issued custom single-item questions to participants, which resulted in \textit{``Good''} representativeness for both visual ($M=3.95$ out of $5$) and mechanical ($M=4.04$ out of $5$) content. Adding the qualitative \textit{Expectation Match} outcomes, both the visual as well as the mechanical parts were quickly able to deliver convincing representations at first shot or some iteration ($88.4\%$ of visuals and $89.8\%$ of mechanics were eventually deemed as fitting the expectation of what the participant had in mind).
The majority of final designs were attributed to be the creators' \textit{own ideas} ($93.5\%$), which further adds to the point that the approach was successful in bringing the participants' ideas to screen. Only $1.45\%$ of cases were eventually attributed to rather model-originated, while the remaining $5.1\%$ were seen as \textit{mixed ideas}. In those cases, participants explicitly revealed that their original idea was improved in a co-creative manner, as they \textit{preferred the generated design to their initial expectation} (7x).
% quote "looked as my bunny!"

\subsubsection*{\textbf{RQ3:}  What are suitable iterative strategies to further close the gap between ideation and generation?}~\\
We deliberately published the whole pipeline as an open-source repository to the target audience (subjects well versed in both generative AI methods as well as visual and mechanical game design) and asked them to configure the approach at will in order to arrive closer at their original idea. As our qualitative codes suggest, a combination of a) adjusting the main Pokémon idea prompt, auxiliary keywords, or, in some cases, parts of the system prompt, and b) ``just iteratively trying again'' was already sufficient to arrive at convincing content ($70.5\%$ combined). Only $6.25\%$ of the participants resorted to changing model parameters (most important: LoRA model strength, CFG guidance, and step size for the visual part; and temperature for the creativity/controllability of the mechanical part), which reportedly was due to the pipeline having had very appropriate parameter configurations in the first place. $0.89\%$ slightly polished their final output manually, e.g., by applying Photoshop or generative image edit models to fix minor visual frictions. $15\%$ indicated a change of their original idea, which in some cases was because the model guided them towards a concept that they preferred (7x), and in others because the model failed to show a good understanding of what they had in mind (10x). $8.93\%$ of subjects simply gave up without trying to adapt, which makes up a large part of the \textit{non-convincing} outputs (see Figure \ref{fig:qual_expectationMatch}).
As $7$ participants mentioned that they would be not qualified to adjust outcomes due to a lack of Pokémon-specific knowledge, we validated our sample by running statistical significance tests between the entire set of subjects and the set without Pokémon knowledge. As shown by $t$-tests for quantitative metrics (as from Figure \ref{fig:quantResults}) and $\chi^2$-tests for qualitative code distributions (as from Figure \ref{fig:qualitativeResults}), including these participants did not significantly change any outcomes ($p>.05$).
All in all, we follow that even though technical understanding and access were available in our subject sample, our proposed pipeline was stable enough to produce mostly convincing outcomes with the least invasive methods of adjustment (prompt refinement, regenerating the same prompt, or following co-creative opportunities) -- which renders it accessible to a wide audience of even novice users, just in the spirit of contemporary, natural-language-centric systems.

\subsubsection*{\textbf{RQ1:} How to unite the rapid advancements of generative AI for content generation of multiple facets of design (visual and mechanical)?}
At this stage, we followed multiple iterations of the pipeline implementation, until we arrived at the approach outlined in Figure \ref{fig:flowchart}. As can be summarized from the just-discussed research questions, this seems to realize a suitable solution to realize player-centric trading card generation that properly matches visual and mechanical design. In short, we highlight the essential building blocks and add anecdotal observations about their ablation:
\begin{itemize}
    \item \textbf{Shared semantic origin for dualistic pipeline.} A minimal representation that maps well to a language representation (mechanics) and visual concepts (artwork) in the deployed models was essential to reach the high measured representativeness. Without this, mismatches between imagery and mechanics are frequently appearing.
    \item \textbf{RAG on existing knowledge base.} Without explicit examples, models would often hallucinate game mechanics. Even if the current results are not entirely safe from hallucinations, it reuses existing abilities, attacks, and core mechanics notably better than without proper RAG.
    \item \textbf{JSON schema output standardization.} Prior to implementing this, language models occasionally delivered incomplete JSON structures, missing values, and empty or structurally underspecified fields.
    \item \textbf{General-purpose tokens and LORAs.} These can largely constrain the artistic direction and style, which however helps to focus on a unified design scope. Based on our experience, missing these allows for a slightly bigger creative range, but at the cost of a clear (visual) direction. As the high satisfaction in visual aesthetics and representativeness indicates, participants were, after all, able to realize their ideas in this slightly narrower but focused scope.
    \item \textbf{Iterative Generation (with clear change directions).} The most promising workflow we explored lies in a human-in-the-loop generative process, which, according to our mixed-methods analyses, can in the majority of cases converge to the initial idea in a few steps. Without clear deltas to adjust or dedicated adjustments to follow, users might lose the connection to the idea they originally had, or are bound to bet on their luck.
\end{itemize}
All in all, there were no critical remarks against the feasibility of our approach (0\%) but 22.4\% even praised the approach (without being asked). Together with the satisfying quantitative and qualitative trends, we conclude that the formerly outlined procedure can be a swift and suitable approach to player-centric content generation.

\section{Limitations \& Future Work}
%Where does BALANCE come into play? (need for balance indentified by )
% We built a game-agnostic combat simulator ... transforms natural language into JSON-structure usable for damage and effect calculation...  make sure the cards that come out of this are balanced (in the context of a reasonably large set of cards, as in "an edition")

While we just listed a series of shortcomings, a fair amount of them can be traced back to the currently known limitations of LLMs themselves (such as hallucinations, limited memory and reasoning, poor understanding of numerical computation, or non-determinism). %As this field is rapidly advancing however, \textbf{empty responses}, \textbf{incomplete fields}, and \textbf{syntax errors} might be problems that right themselves over time (or if deploying a more suitable model than in our current evaluation). Also, these are issues that in theory can be recognized fully automatically, after which a re-prompting iteration procedure might overcome faulty generations practically. Harder issues emerge with semantic limitations, such as \textbf{redundant information} or \textbf{non-strategical mechanics}. 
In upcoming future work, we want to investigate if chain-of-thought prompting can already elevate the reasoning capabilities of mechanical generators such as this one. If this proves to be insufficient, manual teaching and fine-tuning away from unwanted examples might be the most appropriate way to counteract these instances. 
Certainly, innate ethical issues related to LLMs have to be addressed as well, such as environmental concerns in training and execution, the hardly preventable risk of potentially harmful output, and the ongoing debate on copyright or intellectual property when it comes to harnessing large text context, and in this case specifically, custom embeddings.

One of the major criteria of evaluation for procedurally generated content is its \textit{balance}, as generations should be \textit{viable} without strictly dominating (or being dominated by) alternatives. As Summerville et al. point out, this evaluation should be a central part of the generation pipeline already \cite{summerville2018procedural}.
When it comes to \textbf{imbalanced} Pokémon, this is likely a limitation occurring from the current generation pipeline being executed in a vacuum. Traditionally, cards are released in \textit{sets}, where certain cards are deliberately designed to be weaker and more common, whereas others invoke a sense of rarity and superiority by contrast. When adding context about a set of Pokémon cards (in the form of a target distribution, as e.g., Gaussian curves of strength / rarity / types etc.) and potentially automated balance simulations of cards, shortcomings of imbalanced sets could be mitigated to a fair extent \cite{pfau2020dungeons, pfau2022dungeonsII}. In this respect, combos, interactions, and complex strategies between cards are also only feasible to generate if operating on a global set-generation level. In order to tackle this, we already developed a game-agnostic combat simulator that is able to rapidly evaluate the feasibility and balance of generated units \cite{pfau2024damage,pfau2025progression}. It is able to translate the natural language representation of mechanics into functional, logical specifications of game actions and parameters. Since this would exceed the scope and focus of this paper drastically, we decided to omit this evaluation at the submission stage of this paper, and properly reference the findings at publication.

The next major limitation we address in this work is the lack of an evaluation of \textit{play}. While we did assess creators' impressions in contrast to their expectations, they were not deployed into printed or digital play yet. As we see this work as one major stepping stone in a line of research, we emphasize the importance of defining the scope for this paper towards grounding representativeness in the generation process first. Subsequently, we plan to place these generations into well-rounded balance environments and finally evaluate actual feasibility in human play sessions -- necessarily over the longer term to make solid inferences about successful \textit{procedural relatedness} in the end.

In the design of the study for this work, we deliberately decided against a control condition for aesthetics or representativeness. This can be interpreted as a methodological shortcoming, but to the best of our knowledge, no prior approach tackled the generation of player-centric card mechanics yet, which would have rendered any comparison far-fetched. In terms of the visual part of the pipeline, we follow a diffusion-model-agnostic approach anyway, and any existing services that generate trading card or Pokémon artwork are also only built around those\footnote{\url{https://nokemon.eloie.tech/},\url{https://app.scenario.com/?openAppId=wflow_trading-card-maker}}.
Hence, we decided to measure both dimensions against absolute scales, and carefully interpret the outcomes with findings from our qualitative analysis.

% -- (could .. big limitations in text in diffusion models [CITE], 

In addition to the straightaway observable limitations mentioned in Figure \ref{fig:emergentVsFlawed}, there are certainly further subtle challenges to the approach. The current methodology aims towards maximizing \textit{a} plausible completion of a prompt, but not necessarily to a creative or sensible one \cite{maram2023visual}. If anything, the major part of creativity was induced by the participants of this evaluation, whereas the LLM mostly contributed the path of least resistance to the next best solution. This might have resulted in overfitting to certain stats, attacks, and mechanical components. %, e.g. ``Bubble'' as a frequent attack for Water-type Pokémon, or ``Confused'' as a very frequent status effect caused. 
To satisfy a broad creative range, the next iteration of this work will include a method to measure and influence the balance between variability and appropriateness, and evaluate these with the help of established means from procedural content generation.

Looking further into directions for future work, even the initial prompts do not have to be defined by manual inputs, but a set generator could iteratively sample (semi-) random combinations of creature-related words (or portmanteaus, as common for Pokémon), and continue to prompt an LLM to come up with a Pokédex entry for that. In this work, we were more interested in the player-centric generation of their own ideas and the assessment of representativeness thereof. In the next iteration of this work, we are striving to bring the capabilities of this approach full circle.

% As this is the first evaluation of this work, and majorly aimed at demonstrating the feasibility of the approach, the evaluation did not follow quantitative assessments yet, and did not record the participants' opinions on their creations, which would have allowed for more nuanced and consolidated insights (especially regarding the factor of \textit{relatedness}). The resulting interpretation of this evaluation could thus be perceived as too subjective, which is why we aim for a fully-fledged evaluation of the updated pipeline soon.

So far, this pipeline is only capable of generating Pokémon creature cards, disregarding other types (in the case of this TCG, Trainers, Items, Energies, etc.). However, we would argue that a comparable pipeline tailored to the respective card type would be plausible to result in similarly suitable outcomes. 
% %
After all, this initial evaluation could only showcase some feasibility with respect to the (comparably more simple) Pokémon TCG. Technically, the pipeline of Figure \ref{fig:flowchart} can already be deployed on other games of the genre (e.g., for the use case of \textit{Magic: The Gathering}, when replacing the \textit{Pokémon TCG Developer Portal} API by \textit{Scryfall}\footnote{\url{https://scryfall.com/docs/api}} and the \textit{Pokécardmaker} by \textit{Magic Set Editor}), as long as the input prompt follows a suitable card type paradigm. In the long run, we aim to not only cover alternative popular TCGs, but beyond that, to mine and abstract functional data from games beyond TCGs to find mechanics that are game-agnostic and exploit unforeseen mechanical combinations for procedural game generation in general.

\section{Conclusion}
Trading Card Games (TCGs) enable the creation and interplay of rich varieties of playable units, which are well-structured and compact enough to facilitate (possibly limitless) procedural generation of choices. Due to the fact that more complex mechanical logic most often relies on human interpretation of language, however, most prior work was limited to generating parameters procedurally (or creating novel results by iterating combinations/permutations). Since the advent of LLMs, we are looking forward to being able to model and generate mechanical components on a human level, which would open up new pathways for automated generation as well as for game content tailored to individual players and their preferences. In this regard, we constructed a pipeline that aggregates existing TCG data in a structured format, projects them into a custom embedding space to make ground truth knowledge accessible to retrieval-augmented generation, and deploys parallel LLM and diffusion models to turn loosely defined user descriptions into fully-fledged, print-ready cards. Focusing on the case of the Pokémon TCG for an initial proof-of-concept, we carried out an evaluation with $49$ participants that generated $196$ Pokémon cards after their own ideas and preferences, and evaluated these in terms of aesthetics, visual and mechanical representativeness, idea attribution, and possible adaptation methods. Eventually, we report high degrees of visual satisfaction, well-representative artworks \& mechanics, and suitable matching (or exceeding) of expectations. Furthermore, we outline lessons learned, limitations, and reveal trends of desirable \textit{emergent} capabilities. In the long run, we contribute to games user research, to modeling mechanical components of games in general, and we aim to understand overarching patterns across games and how to extend these.

\begin{acks}
% This should be a simple paragraph before the References to thank those individuals and institutions who have supported your work on this article.
We thank all participants. This work is not produced, endorsed, supported, or affiliated with Nintendo, Gamefreak or The Pokémon Company.
\end{acks}

%%
%% The next two lines define the bibliography style to be used, and
%% the bibliography file.
\bibliographystyle{ACM-Reference-Format}
\bibliography{sample-base}

%%
%% If your work has an appendix, this is the place to put it.
%\appendix
%\section{Research Methods}
%\subsection{Part One}

\end{document}